\documentclass[10pt]{article}
\usepackage[utf8]{inputenc}
\usepackage{tcrpreprint}
\usepackage{marvosym}
\usepackage{float}
\usepackage{amsmath}
\usepackage{amssymb}
\usepackage{newunicodechar}
\newunicodechar{π}{\ensuremath{\pi}}
\newunicodechar{₀}{\ensuremath{{}_0}}
\usepackage{array}
\usepackage{tabularx}
\usepackage{booktabs}
\usepackage{colortbl}
\usepackage{graphicx}
\usepackage{microtype}
\usepackage{multirow}
\usepackage{placeins}
\usepackage{wrapfig}
\usepackage{tikz}
\usepackage{url}
\usepackage{xcolor}
\usepackage{hyperref}
\hypersetup{
  pdftitle={Trajectory-Calibrated Merging of Embodied Policies},
  pdfauthor={Wenshuo Chen, Honglei Jin, Le Yang, Jianfei Song, Shaofeng Liang, Lei Wang, Junze Zhu, Kailong Wang, Zhengxi Jin, Haozhe Jia, Honglin Chu, Songning Lai, Xiucheng Wang, Yutao Yue}
}
\usetikzlibrary{arrows.meta,fit,positioning}
\definecolor{TCRBlue}{HTML}{2F5D8A}
\definecolor{TCRLight}{HTML}{EDF3F8}
\definecolor{TCRGray}{HTML}{F3F4F5}
\definecolor{TCRTableBlue}{HTML}{E6F5FF}
\definecolor{TCRTableGray}{HTML}{E0E0E0}
\newcolumntype{Y}{>{\raggedleft\arraybackslash}X}

\newcommand{\method}{Trajectory-Calibrated Regression}
\newcommand{\methodshort}{TCR}

\newcommand{\resulttable}{\small\renewcommand{\arraystretch}{1.12}\setlength{\tabcolsep}{5pt}}
\newcommand{\pizerofive}{\ensuremath{\pi_{0.5}}}

\title{Trajectory-Calibrated Merging\\of Embodied Policies}

\author{\mbox{Wenshuo Chen$^{1,*}$}, \mbox{Honglei Jin$^{1,*\dagger}$}, \mbox{Le Yang$^{2,*}$}, \mbox{Jianfei Song$^{3}$}, \mbox{Shaofeng Liang$^{1}$}, \mbox{Lei Wang$^{4}$}, \mbox{Junze Zhu$^{1}$}, \mbox{Kailong Wang$^{1}$}, \mbox{Zhengxi Jin$^{5}$}, \mbox{Haozhe Jia$^{5}$}, \mbox{Honglin Chu$^{3}$}, \mbox{Songning Lai$^{1}$}, \mbox{Xiucheng Wang$^{6}$}, \mbox{Yutao Yue$^{1,\text{\Letter}}$}}
\affiliation{%
  \mbox{$^{1}$The Hong Kong University of Science and Technology (Guangzhou)} \quad
  \mbox{$^{2}$Sun Yat-sen University} \quad
  \mbox{$^{3}$LimX Dynamics} \quad
  \mbox{$^{4}$Griffith University} \quad
  \mbox{$^{5}$Shandong University} \quad
  \mbox{$^{6}$Xidian University}}
\contribution{$^{*}$Equal contribution \quad $^{\dagger}$Project leader \quad $^{\text{\Letter}}$Corresponding author}
\metadata{Date}{November 2026}
\metadata{Project Page}{\url{https://velixjin.github.io/TCR-Merging/}}
\metadata{Code}{\url{https://github.com/velixjin/TCR-Merging}}

\begin{document}

\begin{abstract}

Robot foundation models are increasingly adapted into task specialists to expand their embodied capabilities.
Yet these capabilities remain fragmented across separate policies, and existing model-merging methods do not reliably preserve them during closed-loop execution:
parameter-space alignment provides only an indirect proxy for behavior, while regression may fail to preserve behavior at states not covered by its calibration inputs. We present \method{} (\methodshort{}), an execution-aware, gradient-free framework that consolidates robot specialists into a single policy.
Across two complementary timescales, TCR collects calibration inputs from expert rollouts spanning different stages of task execution and from intermediate states within iterative action generation.
To account for representation shifts introduced by merging, TCR recomputes each block's input through the already merged blocks. It then uses each expert's computation within the block to form module inputs and fits merged weights to match the experts' linear outputs.
A two-pass calibration procedure further refines the merged policy using separate rollout caches while retaining the frozen specialists as teachers.
The resulting model preserves the original inference graph and requires neither expert routing nor additional gradient training. On LIBERO, TCR merges four \pizerofive{} specialists and achieves \textbf{77.83\%} success across 1,200 evaluation episodes, outperforming FeatCal, RegMean, and direct weight averaging by 9.50, 24.17, and 33.00 percentage points, respectively.
We also find that continued training from a merged checkpoint improves peak success: TCR reaches \textbf{97.75\%} versus 94.00\% when training from the common base.
These results show that calibrating model merging around execution dynamics and evolving internal representations provides an effective path toward preserving diverse embodied skills in a unified policy.

\end{abstract}

\maketitle
\section{Introduction}
\label{sec:introduction}

General-purpose robots need policies that can perform diverse physical tasks.
Vision-language-action models (VLAs) offer a promising foundation~\citep{pmlr-v305-black25a,KimM1-RSS-25}, but demonstration data cover limited tasks and environments, making it difficult to acquire a broad skill set through a single training run.
Training task specialists from a shared foundation model expands coverage, but leaves their capabilities in separate checkpoints.
We seek to consolidate them into one policy.

Model merging can do this without joint training or selecting an expert at inference.
We focus on post-hoc consolidation of independently trained VLA specialists into one static policy, without redesigning the experts or adding a task router.
Parameter-space methods combine weights or align task updates to reduce interference~\citep{pmlr-v162-wortsman22a,yadav2023tiesmerging,stoica2025model}.
But the mapping from weights to actions is nonlinear: reducing conflicts between weights does not ensure that the merged policy preserves expert actions or closed-loop success.
Regression-based methods instead fit merged weights to expert outputs on calibration inputs~\citep{jin2023dataless,nguyen2026regmean}.
Such a fit is only as useful as the inputs it covers.
An action error can change not only the current output but also what the policy sees next; matching a fixed set of expert observations may leave later inputs unconstrained and allow errors to accumulate.

\begin{figure}[t]
\centering
\includegraphics[width=\linewidth]{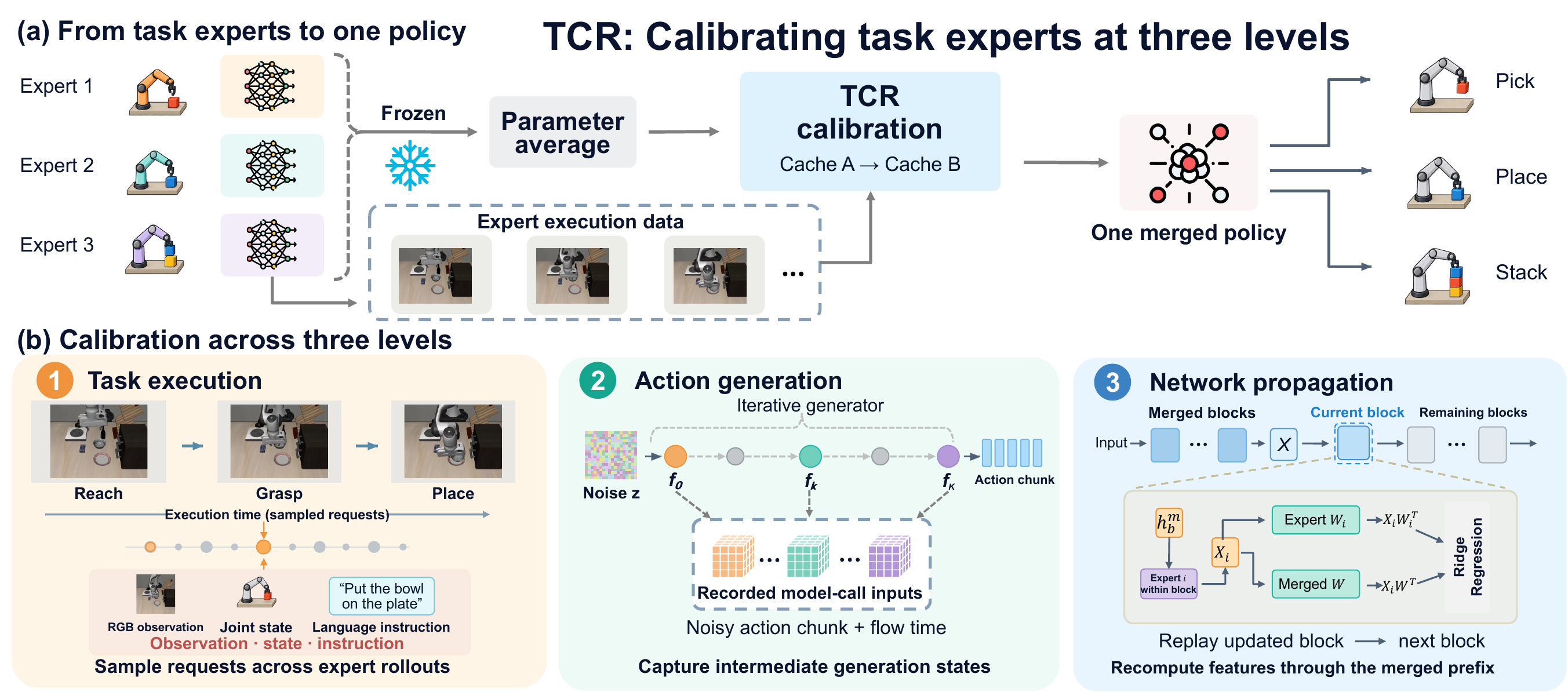}
\caption{Overview of TCR. (a) Two-pass calibration starts from the parameter average and uses separate rollout caches (A/B) from the same frozen experts, yielding one fixed policy. (b) Calibration covers task execution, intermediate action-generation calls, and blockwise replay. Blockwise calibration recomputes merged-prefix features, forms expert-local module inputs for ridge regression, and replays each updated block before proceeding.}
\label{fig:method_overview}
\end{figure}

A slightly misplaced grasp, for example, can change both the camera observation and the robot state at the next decision. Even when the observation is held fixed, the input to an iterative action generator changes across calls as each intermediate prediction updates the action state passed to the next call. Calibration using only the final call can therefore miss states encountered earlier in generation. During merging, a further change occurs within the network. Updating an upstream block can alter the features passed to a downstream block, so a block fitted on the original expert’s features may receive different inputs in the merged policy. The regression fit thus depends on which states are sampled during execution and action generation, and on how the features used to fit each block are computed (see Figure \ref{fig:method_overview}).

We introduce \method{} (\methodshort{}), a gradient-free framework for merging compatible robot experts.
TCR uses frozen experts as teachers and collects two kinds of inputs: observations, robot states, and instructions across expert rollouts, and intermediate call states from each expert's native action generator.
The first covers stages reached after earlier expert actions; the second covers generator inputs, such as noisy actions and flow times, that an observation alone does not specify.
TCR fits adapted perception, conditioning, and action modules to expert outputs on these inputs.
Following blockwise feature replay~\citep{nguyen2026regmean}, it recomputes each block's inputs through the blocks already merged before fitting that block.
Downstream regressions therefore use features produced by the merged prefix rather than relying only on the original expert features.
TCR makes two passes over separate rollout caches: the first updates the parameter average, and the second uses that checkpoint as its prior while still matching the frozen experts.
The second cache adds expert calls not used in the first fit, and the updated prior carries the first pass into the second.
The result is one policy with the original inference graph, no expert router, and no gradient training.

We evaluate TCR on LIBERO~\citep{NEURIPS2023_8c3c6668} using four \pizerofive{} specialists trained for 10,000 steps each.
Across 1,200 episodes, TCR achieves \textbf{77.83\%} success, the highest among the static post-hoc mergers evaluated in this \pizerofive{}-LIBERO setting.
This is 9.50 points above FeatCal (68.33\%), 24.17 above RegMean (53.67\%), and 33.00 above direct weight averaging (44.83\%).
The gains are strongest on Goal and Long, although FeatCal remains ahead on Spatial.
All methods share the same experts and evaluation protocol, while retaining their own calibration budgets.
Ablations test the calibration choices; real-robot and cross-benchmark evaluations test TCR beyond LIBERO.
We also find that merged checkpoints are useful starts for joint training: all four attain higher peak success than common-base training, with TCR highest at \textbf{97.75\%} versus 94.00\% for the base.

Our contributions are:
\begin{itemize}

    \item To our knowledge, we are the first to study calibration
    coverage across both task execution and iterative action generation
    in post-hoc VLA weight merging. Calibration on isolated observations
    can miss states reached later in a task, and even a recorded
    observation leaves unspecified the intermediate inputs used to
    generate its action.

\item We introduce Trajectory-Calibrated Regression (TCR), a
gradient-free method that merges VLA specialists into one router-free
policy. In two passes over separate rollout caches, TCR replays recorded
action-generation inputs through already merged blocks and fits each
subsequent block to frozen expert outputs.

\item Merging four $\pi_{0.5}$ specialists on LIBERO, TCR achieves
77.83\% success over 1,200 episodes, 9.50 percentage points above FeatCal,
the strongest of eight baselines. TCR also leads the evaluated mergers on
LIBERO-PRO, RoboTwin~2.0, and the physical robot (30/40 successes versus
23/40 for FeatCal).

\end{itemize}

\section{Related Work}
\label{sec:related_work}

\subsection{Combining Expert Parameters}
Some merging methods combine parameter updates without explicitly matching model outputs.
Model Soups averages fine-tuned checkpoints; Task Arithmetic scales and combines updates from a common base; TIES and DARE address coordinate conflicts and redundancy~\citep{pmlr-v162-wortsman22a,ilharco2023editing,yadav2023tiesmerging,yu2024language}.
Breadcrumbs, DELLA, and Selective Task Arithmetic refine coordinate selection, while TSV, Iso-CTS, Subspace Boosting, and DC-Merge operate on singular directions or update geometry~\citep{10.1007/978-3-031-73226-3_16,deep2024dellamergingreducinginterferencemodel,bowen2024taskvectorsselectivetask,11092448,marczak2025no,skorobogat2025subspaceboostedmodelmerging,zhang2026dcmergeimprovingmodelmerging}.
KnOTS aligns independently trained LoRA updates, and WUDI and its variants estimate useful input subspaces from weights~\citep{stoica2025model,cheng2025whoever,wei2026closedformspectralregularizationmultitask}.
LoRA-specific and base-anchored methods further constrain update geometry~\citep{Lee_2026_CVPR,ryum2026ctmergingconsensusdirectionstasklevel,zheng2025decoupleorthogonalizedatafreeframework,shi2026pactpreservinganchoredcores}, while generative weight-space models such as Text2Weight synthesize task-specific weights from language and support text-guided fusion~\citep{10.1145/3746027.3755441}.

\subsection{Calibration-Guided Functional Merging}
RegMean and RegMean++ fit merged linear weights to expert outputs, with the latter recomputing features through already merged layers~\citep{jin2023dataless,nguyen2026regmean}.
LOT Merging studies feature drift during model merging and introduces layer-wise task vector fusion to reduce representation mismatch~\citep{NEURIPS2025_fbb4f226}.
ESM uses task-induced layer-wise feature shifts to extract essential subspaces, while activation-guided coefficient methods optimize a more restricted merge parameterization~\citep{Li_2026_CVPR,NEURIPS2025_b395e7bd}.
Fisher merging and AdaMerging additionally use gradient information or optimization on calibration data~\citep{NEURIPS2022_70c26937,yang2024adamerging}.

FeatCal performs closed-form feature calibration in forward order, refreshing features after preceding layers are updated~\citep{gu2026featcalfeaturecalibrationpostmerging}.
TCR builds on regression and sequential feature replay, focusing on calibration inputs collected across robot task execution and action generation.

\subsection{Robot Policies and Specialist Consolidation}
OpenVLA and the $\pi$ family connect pretrained visual-linguistic representations to robot actions~\citep{pmlr-v270-kim25c,BlackK-RSS-25,pmlr-v305-black25a}.
Action chunking predicts several actions per policy request, while diffusion and flow policies, including language-conditioned humanoid controllers such as ECHO and DAJI, generate these chunks through repeated prediction steps~\citep{Zhao-RSS-23,Chi-RSS-23,lipman2023flow,jia2026echoedgecloudhumanoidorchestration,jia2026bodymoveslearninganticipatory}; SmolVLA provides a smaller flow-based implementation~\citep{shukor2025smolvlavisionlanguageactionmodelaffordable}.
OpenVLA-OFT instead decodes continuous actions in parallel~\citep{KimM1-RSS-25}.

Adapting calibration to a different policy architecture requires matching its actual inference computation.

\paragraph{Combining robot skills.}
FLEET-MERGE studies policy merging for robot fleets by aligning separately trained policies under parameter permutation symmetries~\citep{ICLR2024_e28b3369}.
MergeVLA explores robot skill consolidation by modifying expert construction with sparse masks to encourage mergeability~\citep{Fu_2026_CVPR}.
OSRM studies LoRA mergeability in language models through structured adapter subspaces~\citep{zhang-zhou-2025-unraveling}.
TCR instead starts with existing compatible experts and fits merged weight matrices using inputs recorded during task execution.
CapVector transfers capability vectors for subsequent adaptation, while VLA-Pro retrieves and dynamically fuses adapter memories at inference time~\citep{song2026capvectorlearningtransferablecapability,si2026vlaprocrosstaskproceduralmemory}.
RETAIN interpolates generalist and fine-tuned VLA parameters and uses the merged checkpoint to initialize adaptation to a subsequent robot task~\citep{yadav2026robust}.
In language models, Mashup Learning merges historical checkpoints into an initialization for further fine-tuning~\citep{locicerovaina2026mashup}.
Our continuation study instead jointly trains the original source tasks after merging four specialists under matched updates.
ReVLA restores visual generalization after adaptation through representation recovery~\citep{11128823}.

The dependence of future inputs on current predictions is also central to imitation learning~\citep{pmlr-v15-ross11a}.
Policy distillation further studies transferring behaviors from expert policies by matching student predictions to teacher outputs on expert-generated data~\citep{rusu2016iclr-policy}.
TCR does not perform dataset-aggregation policy learning; instead, it also uses expert-generated inputs and fits the merged weights through fixed regression rather than training a student policy.

\section{Trajectory-Calibrated Regression}
\label{sec:method}

TCR collects frozen expert inputs across task execution and action generation, then fits modules with merged-prefix replay (Figure~\ref{fig:method_overview}).
We use $\Theta$ for the full set of policy parameters and $W$
for the weight matrix of a single linear module within that policy.
We denote finite calibration caches by $D$ and input distributions
by $\mu$. The superscript $M$ denotes merged quantities.

\subsection{Experts, Output Policy, and Calibration Objective}
Given $M$ experts $\pi_i\equiv\pi_{\Theta_i}$ trained independently from the same initial model, our goal is one policy that retains their skills without selecting an expert at inference.
Experts share an architecture, tokenizer, observation and action formats, and compatible normalization; the merged policy retains language conditioning.

Matching expert outputs provides a general objective for functional merging:
\begin{equation}
 \Theta^\star=\arg\min_\Theta\sum_{i=1}^{M}
 \mathbb{E}_{x\sim\mu_i}\,
 \mathcal L\bigl(f_\Theta(x),f_{\Theta_i}(x)\bigr).
 \label{eq:functional_merging}
\end{equation}
Here $\mu_i$ is the matching-input distribution and $\mathcal L$ is the output discrepancy.
The $x$ in Eq.~\eqref{eq:functional_merging} denotes a generic matching input; below, we distinguish request-level $x_r$ from call-level $\tilde x_{r,f}$.

During execution, a robot's actions affect what it observes next,
and each intermediate action prediction affects the input to the next
generator call. The inputs encountered at deployment therefore depend on
the policy being executed. TCR approximates the matching objective by
fitting local linear outputs on a finite set of inputs recorded during
frozen expert rollouts.

For a given linear module, let $W_i$ denote its weight matrix
in expert $i$. We fit full linear weight matrices across adapted perception, conditioning, and action modules, including directly trained interfaces and biases.
Identical parameters remain unchanged; other adapted tensors follow explicit fallback rules (Appendix~\ref{app:materialization}).

\subsection{Collecting Calibration Inputs}
\paragraph{Across task execution.}
To include inputs reached after earlier expert actions, we sample from complete expert episodes.
Write $x_r=(o_r,s_r,\ell)$ for the visual observation, proprioceptive state, and task instruction at request $r$.
Let $\mathcal H_r$ contain retained policy history.
Expert $i$'s native generator produces the action chunk $\mathbf a_r$ and updates this history:
\begin{equation}
 (\mathbf a_r,\mathcal H_{r+1})=\mathcal G_{\Theta_i}(x_r,\mathcal H_r;\xi_r),
 \qquad
 x_{r+1}\sim\mathcal P_{\mathrm{env}}\bigl(\,\cdot\mid x_r,\operatorname{Exec}(\mathbf a_r)\bigr).
 \label{eq:policy_loop}
\end{equation}
Here $\xi_r$ denotes sampling randomness, and $\operatorname{Exec}$ selects the actions actually executed from the predicted chunk.
Here $x_{r+1}=(o_{r+1},s_{r+1},\ell)$, with $\ell$ fixed within the episode; the transition notation abbreviates the underlying environment dynamics, not a Markov assumption about observations.
We sample requests throughout each expert's assigned-task episodes, retaining all inputs needed for replay.

\paragraph{Within action generation.}
An observation alone does not specify every input used by an iterative action generator.
A calibration record $\tilde x_{r,f}=(x_r,\mathcal H_r,\mathbf u_{r,f},f)$ contains the inputs needed to reconstruct an actual model call.
The internal state $\mathbf u_{r,f}$ and call index $f$ exist only where the generator uses them.
Within each sampled request, we retain selected calls from the expert's native generation procedure.

For \pizerofive{}, $\mathbf u_{r,f}$ includes a noisy action chunk and its flow time.
Other generators can have different inputs: a parallel action decoder has no denoising coordinate, while a video-generating policy may require additional visual latents and history.
Appendix~\ref{app:backbone_contract} outlines the checks needed before extending TCR beyond the evaluated \pizerofive{} model.

\paragraph{Sampling and budget.}
Let $\mu_i^E$ denote the call-input distribution induced by expert $\pi_i$ under the sampling rule, and $\mathcal I_i^E$ its retained request-call index pairs.
For $\pi_{0.5}$, a recorded call pairs the policy request and its
history with the noisy action chunk and flow time at that generator step.
To draw a call input from the mixture $\mu^E$, we first choose one of the
$M$ experts uniformly and then draw a call input from that expert's
distribution $\mu_i^E$.
\begin{equation}
 \mu^E=\frac{1}{M}\sum_{i=1}^M\mu_i^E,
 \qquad \mathcal D_i^E=\{\tilde x^{(i)}_{r,f}:(r,f)\in\mathcal I_i^E\}.
 \label{eq:trace_measure}
\end{equation}
We divide the samples by task, episode, position within the episode, and action generation step.
We do not keep only successful episodes.
For iterative models, each stored observation remains paired with the intermediate states generated from it.

The calibration budget counts feature rows $n_{i,b,j}$ actually used per expert and module, rather than stored calls, whose sequence lengths and branches may differ.
We allocate row quotas across sample groups without duplicating scarce samples (Appendix~\ref{app:calibration_recipes}).
Temporal order determines input collection, but shuffling the resulting input-teacher pairs leaves the regression objective unchanged.

\subsection{Fitting the Merged Policy}
\paragraph{Recompute features before fitting each block.}
Earlier merged layers can produce different features from the experts.
Following sequential replay~\citep{nguyen2026regmean}, we visit blocks in forward order, recomputing each block's inputs through the already merged prefix.
Let $h_b^{\mathrm M}(\tilde x)$ denote this input to block $b$ for record $\tilde x$.
Within the current block, we still use each expert's computations before the linear module being fitted; denote these computations by $\phi_{i,b,j}$.
Its regression inputs are
\begin{equation}
 X_{i,b,j}=\operatorname{Rows}
 \left\{\phi_{i,b,j}\bigl(h_b^{\mathrm M}(\tilde x)\bigr):\tilde x\in\mathcal D_i^E\right\}.
 \label{eq:replay_inputs}
\end{equation}
We solve the block's linear modules on these fixed features, write their merged weights, and recompute the block outputs before proceeding.

When the network has several branches, a component is processed after the components that supply its inputs.
Conditioning representations, cross-attention keys and values, and time embeddings are refreshed when their producers change.
The evaluated \pizerofive{} implementation covers 418 linear modules and 422 adapted tensors (Appendix~\ref{app:replay_structure}); these counts do not transfer to other architectures.
Before calibration, each replay implementation should reproduce the original forward pass at identical weights and inputs.
Stored generation states and execution history remain fixed during replay; neither is regenerated by the merged policy.

\paragraph{Match the experts' linear outputs.}
For each module, $P$ is its fixed weight in the current pass's prior checkpoint: the expert mean in pass one, then the first-pass checkpoint.
Suppressing block and module indices, $W_i,P\in\mathbb R^{d_{\mathrm{out}}\times d_{\mathrm{in}}}$ and $X_i\in\mathbb R^{n_i\times d_{\mathrm{in}}}$.
Let $\omega_i$ be the expert loss mass, with $\sum_i\omega_i=1$; dimensions and masses are module-specific.
We fit the correction to preserve expert outputs on $X_i$, while ridge regularization limits changes from $P$:
\begin{equation}
 C^\star=\arg\min_C
 \sum_i\frac{\omega_i}{n_i}
 \left\|X_i(P+C)^\top-X_iW_i^\top\right\|_F^2
 +\lambda\|C\|_F^2,
 \qquad W^{\mathrm M}=P+C^\star .
 \label{eq:tcr_objective}
\end{equation}
Define $\Gamma_i=(\omega_i/n_i)X_i^\top X_i$ and $\Gamma=\sum_i\Gamma_i$.
Following the regression-based merging principle of RegMean~\citep{jin2023dataless} , we use the following weighted prior-centered ridge solution:
\begin{equation}
 W^{\mathrm M}=\left(\sum_i W_i\Gamma_i+\lambda P\right)(\Gamma+\lambda I)^{-1}.
 \label{eq:tcr_solve}
\end{equation}
We use linear solves rather than explicitly forming the inverse.
Positive ridge makes the system well-defined even with dependent rows.
For any direction $\mathbf v$ with $X_i\mathbf v=0$ for every expert, $C^\star\mathbf v=0$: the prior is retained.
Thus regression retains the prior in directions not covered by calibration features.
The equivalent smaller system, bias handling, ridge scaling, and post-solve safeguard are given in Appendix~\ref{app:method}.

\paragraph{Weighting the experts.}
In the first pass, TCR weights each expert by how closely the prior already matches its linear outputs:
\begin{equation}
 e_i^{\mathrm{prior}}=\frac{\|X_i(P-W_i)^\top\|_F^2}{n_i d_{\mathrm{out}}},
 \qquad
 \omega_i^{\mathrm{rel}}=\frac{1/\max(e_i^{\mathrm{prior}},\epsilon)}
 {\sum_{i'=1}^{M}1/\max(e_{i'}^{\mathrm{prior}},\epsilon)}.
 \label{eq:expert_mass}
\end{equation}
These weights are computed before solving each module and remain fixed during that solve.
Without the numerical floor, the data term is proportional to
$\sum_i E_i(C)/E_i(0)$, where $E_i(C)$ is the unnormalized squared residual.
Under this normalization, doubling the output error of a module
incurs the same increase in the data term for any expert, regardless of
its initial error. We use this criterion in the first pass so that the
fit accounts for the cost of degrading expert outputs that the initial
weight average already matches closely.
The second pass instead assigns uniform expert masses, $\omega_i^{\mathrm{uni}}=1/M$.
Uniform masses give equal weight to expert losses, not equal parameter averaging.

\subsection{Two Calibration Passes, One Deployed Policy}
Let $\mathcal S(\Theta,\mathcal D;\mathrm{rule})$ denote one blockwise pass with fixed checkpoint prior $\Theta$, expert cache $\mathcal D$, and weighting rule.
Using separate caches $\mathcal D_A^E$ and $\mathcal D_B^E$ from the same frozen experts, TCR starts from $\bar\Theta=M^{-1}\sum_i\Theta_i$ and computes
\begin{equation}
 \Theta^{(1)}=\mathcal S(\bar\Theta,\mathcal D_A^E;\mathrm{rel}),
 \qquad
 \Theta^{\mathrm{TCR}}=\mathcal S(\Theta^{(1)},\mathcal D_B^E;\mathrm{uni}).
 \label{eq:two_pass_tcr}
\end{equation}
Both passes match the frozen experts using forward execution and linear solves; the second fits cache B with $\Theta^{(1)}$ as its prior.
Only $\Theta^{\mathrm{TCR}}$ is deployed, retaining the original graph without expert routing or online recalibration.
Per-pass sampling and budgets are given in Appendix~\ref{app:calibration_recipes}.
Because the second pass uses $\Theta^{(1)}$ as its prior, the two passes are successive proximal updates rather than one pooled fit.
In a simplified model with fixed features, two successive updates contract the initial error more than a single pass on the pooled caches (Appendix~\ref{app:sequential_prior}); empirically, further passes help mainly when they use fresh caches (Figure~\ref{fig:calibration_passes}).

\section{Experiments}
\label{sec:experiments}
We test skill retention, calibration choices, action agreement, real-robot behavior, and cross-benchmark performance. Supplementary controls separate execution coverage, repeated calibration, and data budget.

\subsection{Evaluation Setting}
LIBERO Spatial, Object, Goal, and Long define four ten-task specialists~\citep{NEURIPS2023_8c3c6668}.
The main LIBERO comparison uses the same four frozen \pizerofive{} experts, trained from a common initialization for 10,000 steps each.
Preprocessing, action representation, chunk execution, and serving mode are fixed across mergers within each evaluation setting.
The main LIBERO comparisons report success as the mean $\pm$ sample standard deviation across three evaluations of 400 episodes each (ten per task), with shared task and initial-state assignments across methods.
Appendix~\ref{app:protocol} details data splits, evaluation repeats, and generation-noise handling.

\subsection{Skill Retention after Merging}
\label{sec:main_results}
Can one fixed policy retain several specialists' skills? We compare average and suite-level retention against own-suite experts.

\paragraph{Simulation.}
Table~\ref{tab:main_baselines} compares eight baselines on the common expert bank and protocol; Appendix~\ref{app:baseline_configs} details their recipes and calibration budgets.

\begin{table}[t]
\centering
\caption{LIBERO success rates (\%) for four $\pi_{0.5}$ specialists, each trained for 10k steps, and their merged policies. We report the mean $\pm$ sample standard deviation across three runs of 400 episodes each. For TCR, the three runs evaluate three independently constructed models with separate calibration caches; for every other method, they evaluate the same checkpoint three times. Blue highlights TCR. In each column, bold marks the specialist reference and the best merged policy. Appendix~\ref{app:baseline_configs} details the baseline adaptations and numerical settings.}
\label{tab:main_baselines}
\resulttable
\newcommand{\meanstd}[2]{#1^{\scriptstyle\pm #2}}
\begin{tabularx}{\linewidth}{lYYYYY}
\toprule
\textbf{Method} & \textbf{Spatial} & \textbf{Object} & \textbf{Goal} & \textbf{Long} & \textbf{Average} $\boldsymbol{\uparrow}$ \\
\midrule
\rowcolor{TCRTableGray}
Experts$^{\mathrm{ref}}$ & $\meanstd{\mathbf{92.00}}{1.00}$ & $\meanstd{\mathbf{99.33}}{1.15}$ & $\meanstd{\mathbf{97.33}}{2.52}$ & $\meanstd{\mathbf{97.00}}{2.00}$ & $\meanstd{\mathbf{96.42}}{0.76}$ \\
\midrule
Model Soups & $\meanstd{68.33}{3.51}$ & $\meanstd{54.33}{2.52}$ & $\meanstd{46.33}{1.15}$ & $\meanstd{10.33}{1.15}$ & $\meanstd{44.83}{1.81}$ \\
Task Arithmetic & $\meanstd{68.67}{1.15}$ & $\meanstd{56.00}{2.65}$ & $\meanstd{42.00}{1.73}$ & $\meanstd{17.33}{1.53}$ & $\meanstd{46.00}{0.75}$ \\
TIES-Merging & $\meanstd{65.00}{5.00}$ & $\meanstd{54.33}{0.58}$ & $\meanstd{37.33}{1.53}$ & $\meanstd{20.33}{2.52}$ & $\meanstd{44.25}{1.30}$ \\
WUDI-Merging & $\meanstd{67.00}{1.00}$ & $\meanstd{60.67}{2.08}$ & $\meanstd{38.00}{2.00}$ & $\meanstd{18.33}{2.52}$ & $\meanstd{46.00}{0.43}$ \\
KnOTS-TIES & $\meanstd{69.33}{3.21}$ & $\meanstd{69.33}{5.03}$ & $\meanstd{44.67}{2.52}$ & $\meanstd{6.67}{1.15}$ & $\meanstd{47.50}{2.63}$ \\
RegMean & $\meanstd{71.33}{3.51}$ & $\meanstd{81.33}{0.58}$ & $\meanstd{46.67}{2.52}$ & $\meanstd{15.33}{2.08}$ & $\meanstd{53.67}{0.63}$ \\
RegMean++ & $\meanstd{76.00}{3.00}$ & $\meanstd{86.00}{1.00}$ & $\meanstd{50.00}{1.00}$ & $\meanstd{20.67}{2.08}$ & $\meanstd{58.17}{1.04}$ \\
FeatCal & $\meanstd{\mathbf{83.00}}{2.65}$ & $\meanstd{89.00}{1.00}$ & $\meanstd{53.33}{1.53}$ & $\meanstd{48.00}{3.61}$ & $\meanstd{68.33}{0.76}$ \\
\midrule
\rowcolor{TCRTableBlue}
\textbf{\methodshort{} (ours)} & $\meanstd{80.00}{3.00}$ & $\meanstd{\mathbf{94.00}}{2.65}$ & $\meanstd{\mathbf{74.00}}{6.24}$ & $\meanstd{\mathbf{63.33}}{3.21}$ & $\meanstd{\mathbf{77.83}}{2.52}$ \\
\bottomrule
\end{tabularx}
\par\smallskip
\begin{minipage}{\linewidth}

\end{minipage}
\end{table}

TCR obtains 934/1200 successes (77.83$\pm$2.52\%), exceeding FeatCal, the strongest baseline, by 9.50 percentage points (820/1200; 68.33$\pm$0.76\%).
The gain is concentrated in Goal and Long: TCR exceeds FeatCal by 20.67 and 15.33 percentage points, respectively, versus 5.00 on Object and a 3.00-point deficit on Spatial.
Long also exposes a large gain over direct averaging (63.33\% versus 10.33\%); the specialist reference is 97.00\%.
Section~\ref{sec:mechanism} tests observation source, generation inputs, and network features directly.

Appendix~\ref{app:qualitative_rollouts} shows matched execution examples, including a drawer-opening task where TCR succeeds and Soup, TIES, and FeatCal fail.

\paragraph{Real robot.}
\label{sec:real_robot}
The hardware study uses \pizerofive{} task specialists on a PiperX robot, trained from a common base, and shared calibration traces from their physical executions.
Table~\ref{tab:real_robot} compares each merger (Model Soups, TIES, RegMean++, FeatCal, and TCR) against own-task specialists across 20 trials per task (40 per method).
Each trial has a 60\,s execution limit; timeouts and assisted completion do not count as autonomous success.
Appendix~\ref{app:real_robot} gives collection and safety procedures.

\begin{table}[ht]
\centering
\caption{Real-robot success (20 trials/task, 60\,s/trial) and offline action RMSE. Task~1 is plate placement; Task~2 is stacking; Avg. weights them equally. RMSE pools action elements against the corresponding expert, not task completion (Appendix~\ref{app:real_robot}). Each expert is evaluated on its own task, so its RMSE against itself is zero.
Each merged policy uses one checkpoint for both tasks.
Bold indicates the highest success rate within each group and the lowest RMSE among merged policies.}
\label{tab:real_robot}
\resulttable
\begin{tabularx}{\linewidth}{lYYYYY}
\toprule
& \multicolumn{3}{c}{Real-robot success (\%, $\uparrow$)} & \multicolumn{2}{c}{Offline action RMSE ($\downarrow$)} \\
\cmidrule(lr){2-4}\cmidrule(lr){5-6}
Method & Task 1 & Task 2 & Avg. & Plate placement & Stacking \\
\midrule
\rowcolor{TCRTableGray}
Experts$^{\mathrm{ref}}$ & \textbf{85.0} & \textbf{90.0} & \textbf{87.5} & 0.0000 & 0.0000 \\
\midrule
Model Soups & 15.0 & 5.0 & 10.0 & 0.0985 & 0.0648 \\
TIES & 20.0 & 10.0 & 15.0 & 0.1063 & 0.0826 \\
RegMean++ & 45.0 & 55.0 & 50.0 & 0.0738 & 0.0563 \\
FeatCal & 70.0 & 45.0 & 57.5 & 0.1084 & 0.0693 \\
\midrule
\rowcolor{TCRTableBlue}
\textbf{\methodshort{}} & \textbf{75.0} & \textbf{75.0} & \textbf{75.0} & \textbf{0.0471} & \textbf{0.0486} \\
\bottomrule
\end{tabularx}
\par\smallskip
\end{table}

The own-task specialists succeed in 35/40 trials overall (87.5\%), and TCR succeeds in 30/40 (75.0\%).
Compared with FeatCal's 23/40 (57.5\%), TCR's advantage is larger on stacking (75\% versus 45\%) than plate placement (75\% versus 70\%).
TCR also has the lowest expert-relative action RMSE among mergers on both tasks, but action agreement does not determine the success ranking: on plate placement, FeatCal succeeds more often than Soup (70\% versus 15\%) despite higher RMSE (0.1084 versus 0.0985).
Errors can affect subsequent interactions differently; RMSE alone does not capture this distinction.

\subsection{Which Calibration Choices Matter?}
\label{sec:mechanism}
Actions shape observations, generator calls shape later inputs, and merged blocks shape downstream features. Table~\ref{tab:mechanism_ablation} replaces one calibration choice at a time, in both passes, with three 400-episode evaluations per variant.

\paragraph{Task execution: where do observations come from?}
\begin{table}[t]
\centering
\begin{minipage}{0.66\linewidth}
\centering
\caption{Calibration ablations on \pizerofive{}-LIBERO. Success (\%) is mean $\pm$ standard deviation over three 400-episode evaluations.}
\label{tab:mechanism_ablation}
\small\renewcommand{\arraystretch}{1.15}\setlength{\tabcolsep}{5pt}
\begin{tabularx}{\linewidth}{lYYY}
\toprule
\textbf{Variant} & \textbf{MSE} $\downarrow$ & \textbf{Cos.} $\uparrow$ & \textbf{Success} $\uparrow$ \\
\midrule
\rowcolor{TCRTableBlue}
\textbf{Full} & $\mathbf{0.0488}^{\scriptstyle\pm 0.0011}$ & $\mathbf{0.8620}^{\scriptstyle\pm 0.0017}$ & $\mathbf{77.83}^{\scriptstyle\pm 2.52}$ \\
No rollouts & $0.0513^{\scriptstyle\pm 0.0008}$ & $0.8545^{\scriptstyle\pm 0.0015}$ & $73.00^{\scriptstyle\pm 1.52}$ \\
Final call & $0.0516^{\scriptstyle\pm 0.0014}$ & $0.8571^{\scriptstyle\pm 0.0022}$ & $73.17^{\scriptstyle\pm 2.31}$ \\
No replay & $0.0655^{\scriptstyle\pm 0.0008}$ & $0.8302^{\scriptstyle\pm 0.0018}$ & $64.50^{\scriptstyle\pm 1.52}$ \\
\bottomrule
\end{tabularx}
\end{minipage}
\end{table}

We substitute demonstration observations, proprioception, and instructions for rollout inputs, retaining frozen teachers, their generator, and the regression rule. Expert outputs are recomputed on these inputs; demonstration actions are not regression targets. With task allocations, requests, calls, and rows matched, success falls from 77.83\% to 73.00\%. Every repeat and suite average declines, by 2.33 points on Spatial to 7.33 on Long (Table~\ref{tab:three_level_suite_results}). Rollout-derived observations thus improve retention across all four suites in this comparison; Appendix~\ref{app:detailed_success} reports task-cluster uncertainty, and Appendix~\ref{app:recipe_controls} controls for early rollout coverage.

\paragraph{Action generation: which model calls are calibrated?}
We keep the same expert requests but retain only the final generator call. For \pizerofive{}, calls differ in noisy action state and flow time even at a fixed observation. With per-call sampling unchanged, first-pass rows fall from 4,441,200 to 1,480,400; second-pass rows remain 1,331,600. Success falls to $73.17\pm2.31\%$, supporting intermediate-call calibration in this recipe.

\paragraph{Network propagation: whose features are fitted?}
Replacing merged-prefix features with expert features, while keeping records, calls, regression, and parameter scope fixed, yields $64.50\pm1.52\%$, the lowest variant. This supports merged-prefix replay within the tested construction; the strategy follows prior work~\citep{nguyen2026regmean}.

All variants retain first-pass relative-error weighting and second-pass uniform weighting from each arm's first-pass checkpoint. Together, the replacements support these three choices within the evaluated recipe. Appendix~\ref{app:ablation_protocol} details numerical controls and supplementary studies.

\textbf{Does action agreement accompany skill retention?} On 400 requests from 80 held-out expert episodes, each policy runs its own action generator from the same observations, history, and initial noise as its expert. We compare the six normalized continuous action coordinates over the ten steps used for execution. Among the calibration variants, full TCR has the lowest action MSE (0.0488), the highest cosine similarity (0.8620), and the highest closed-loop success (77.83\%). Across nine merging methods, the Spearman correlation between lower MSE and higher success is 0.86 over all four LIBERO suites. Holding this offline ranking fixed, the correlation is 0.95 for mean success on Spatial, Object, and Goal, but only 0.52 for Long.

For example, KnOTS-TIES has lower overall MSE than TIES-Merging (0.0980 vs. 0.1169) and higher mean success on the first three suites (61.11\% vs. 52.22\%), yet succeeds less often on Long (6.67\% vs. 20.33\%). Offline MSE measures agreement at states visited by the expert. In a multistage task, a single missed grasp can prevent later stages from succeeding, and the offline requests do not test how the policy responds to the state created by its own mistake. Action agreement therefore helps diagnose skill retention, with a weaker correspondence to long-horizon success in this comparison.
\begin{figure}[H]
    \centering
    \includegraphics[width=\linewidth]{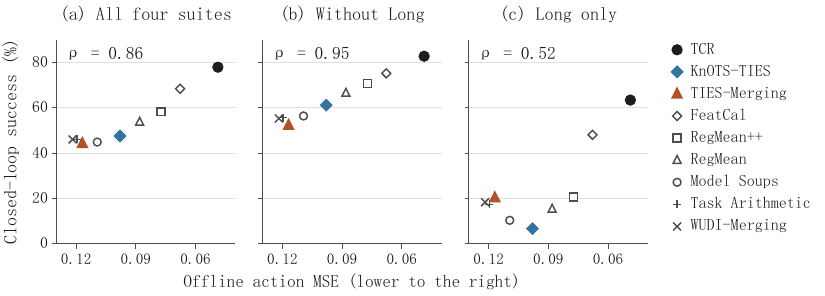}
    \caption{ Action agreement and closed-loop success on LIBERO.
    All panels use the same four-suite action MSE on 400 held-out expert requests.
    Success is averaged over (a) all four suites, (b) Spatial, Object, and Goal,
    or (c) Long alone. Spearman's $\rho$ is computed between $-\mathrm{MSE}$
    and success across nine merging methods.}
    \label{fig:action-agreement-by-suite}
\end{figure}

\subsection{Evaluation Across Benchmarks}
\label{sec:cross_setting}
\begin{table}[t]
\centering
\begin{minipage}{0.58\linewidth}
\centering
\caption{Cross-benchmark success (\%).}
\label{tab:cross_benchmark}
\small\renewcommand{\arraystretch}{1.00}\setlength{\tabcolsep}{5pt}
\begin{tabularx}{\linewidth}{@{}lYY@{}}
\toprule
\textbf{Method} & \textbf{LIBERO-PRO} $\boldsymbol{\uparrow}$ & \textbf{RoboTwin 2.0} $\boldsymbol{\uparrow}$ \\
\midrule
\rowcolor{TCRTableGray}
Experts$^{\mathrm{ref}}$ & 50.75 & 57.78 \\
\midrule
Model Soups & 28.81 & 12.22 \\
TIES & 27.13 & 6.11 \\
RegMean++ & 35.38 & 13.33 \\
FeatCal & 41.38 & 0.00 \\
\midrule
\rowcolor{TCRTableBlue}
\textbf{\methodshort{} (ours)} & \textbf{43.19} & \textbf{15.56} \\
\bottomrule
\end{tabularx}
\end{minipage}
\end{table}

\paragraph{LIBERO-PRO.}
We evaluate frozen LIBERO checkpoints under position, object, language, and changed-goal perturbations, without recalibration or selection on PRO~\citep{zhou2026liberoprorobustfairevaluation}.
TCR obtains the highest average among the evaluated mergers (43.19\%); the expert reference is 50.75\%. Appendix~\ref{app:benchmark_task_reporting} reports results by perturbation type.

\paragraph{RoboTwin 2.0.}
Table~\ref{tab:cross_benchmark} summarizes \pizerofive{} on LIBERO-PRO and RoboTwin 2.0.
Unlike PRO's frozen-policy evaluation, RoboTwin uses separately trained bimanual experts for the Coordination, Receptacle, and Precision task groups~\citep{chen2025robotwin20scalabledata}.
It tests consolidation of a new expert bank on its source tasks, not zero-shot transfer of the LIBERO policy.

On the separately trained RoboTwin expert bank, TCR leads the evaluated mergers at 15.56\%, versus 13.33\% for RegMean++; the specialist reference reaches 57.78\%.
Task-group results and PRO perturbation breakdowns appear in Appendix~\ref{app:benchmark_task_reporting}.
\label{sec:scaling}
Appendix~\ref{app:expert_count} varies the expert set: TCR leads at two and four experts, whereas FeatCal is higher at three; task coverage grows with the set.
Appendix~\ref{app:cost} accounts separately for calibration, fusion, and deployment costs.

\subsection{Further Training from Merged Initializations}
\label{sec:continued}
Can a merged checkpoint improve joint training after its immediate zero-update advantage? We train the common base, Model Soups, RegMean++, FeatCal, and TCR on the same source tasks with matched data order, optimizer, schedule, batch size, and update budget, resetting optimizer states. Table~\ref{tab:continued} reports fixed-update scores and peaks on a shared evaluation grid; Appendix~\ref{app:downstream} gives the full curves (Figure~\ref{fig:continued_training}) and per-suite results.

The benefit persists during training: at 5,000 updates TCR reaches 96.75\% versus 94.00\% for the common base, and its 0--5,000-update nAUC is 92.83\% versus 73.84\%. All four merged initializations also surpass the base's best observed success (94.00\%), with peaks from 95.50\% to 97.75\%. TCR has the highest observed peak, 3.75 points above the base and 0.50 points above FeatCal. Thus, in this matched setting, merging is useful not only for immediate deployment but also as an initialization from which joint training attains higher observed success.

\begin{table}[h]
\centering
\caption{Further LIBERO training (\% success; 400 episodes/checkpoint). Peak and its step use the extended grid; nAUC uses the 0--5,000-update window.}
\label{tab:continued}
\resulttable
\begin{tabularx}{\linewidth}{lYYYYY}
\toprule
Initialization $m$ & $S_m(0)$ & $S_m(5{,}000)$ & $S_m(10{,}000)$ & $\operatorname{nAUC}_m^{0:5k}$ & Peak (step) \\
\midrule
Common base & 0.25 & 94.00 & 92.00 & 73.84 & 94.00 (5,000) \\
Model Soups & 47.50 & 95.25 & 93.00 & 88.36 & 95.50 (6,500) \\
RegMean++ & 58.75 & 94.75 & 94.50 & 88.50 & 95.50 (5,500) \\
FeatCal & 65.50 & 95.25 & \textbf{96.25} & 92.09 & 97.25 (8,000) \\
\rowcolor{TCRTableBlue}
\methodshort{} & \textbf{78.75} & \textbf{96.75} & \textbf{96.25} & \textbf{92.83} & \textbf{97.75} (6,500) \\
\bottomrule
\end{tabularx}
\end{table}

\FloatBarrier

\section{Discussion and Limitations}
\label{sec:discussion}

Execution-aware calibration improves skill retention: rollout observations, intermediate states, and merged-prefix replay all increase success. Embodied merging should therefore consider execution dynamics and internal features alongside parameter compatibility, with performance gains tied to calibration coverage.

TCR yields a single static policy by replaying native inference states. Spanning LIBERO, LIBERO-PRO, RoboTwin 2.0, and a physical robot, it achieves the highest aggregate success among evaluated mergers. It preserves the original inference graph without deployment-time routing or training, requiring only one-time offline calibration.

\textbf{Limitations and Methodological Bounds.} A single static set of weights has limited capacity for conflicting skills: when experts use shared feature directions differently, no shared linear map can match all of them (Appendix~\ref{app:conflict_bound}), and retention decreases smoothly from 91.2\% to 76.1\% as two to four experts are merged (Appendix~\ref{app:limits_count}); larger expert sets remain untested. TCR also requires white-box access to specialists with compatible architectures and observation and action interfaces, so merging across embodiments is left to future work.

\textbf{Systematic Future Work.} These boundaries expose broader challenges for embodied consolidation, inspiring three research directions: 
(1) \textit{From Expert to Recovery-Aware Manifolds:} Gradient-free merging currently relies on manifolds covered by routine expert rollouts. In closed-loop dynamics, minor contact deviations trigger cascading errors, pushing merged policies into out-of-distribution states. This is consistent with the weaker correlation between offline action MSE and success on Long-horizon tasks ($\rho=0.52$; Sec.~\ref{sec:mechanism}). Future research should constrain the ``recovery manifold'' by introducing active perturbations (e.g., DAgger-style sampling) or deviation states synthesized by world models into calibration, shifting from mimicking success to calibrating recovery. 
(2) \textit{Black-Box and Cross-Embodiment Merging:} Extending merging to API policies or heterogeneous robots, such as language-conditioned humanoids~\citep{jia2026bodymoveslearninganticipatory,jia2026echoedgecloudhumanoidorchestration}, requires breaking dependence on homologous weights. Future frameworks could explore representation distillation, mapping disparate behaviors into a shared latent space or student models before trajectory-calibrated fusion, bridging independent architectures. 
(3) \textit{Continual Merging:} TCR currently recalibrates with all experts whenever a specialist is added. Because each pass already uses the previous result as its prior, a natural extension is to add specialists incrementally, calibrating on the new expert and a small replay set of earlier ones; whether this preserves earlier skills remains to be tested.

\section{Conclusion}
\label{sec:conclusion}
\method{} records inputs across task execution and action generation, then fits merged weights using features recomputed through the partially merged policy.
Its output is one policy with fixed weights and no additional expert router.
TCR reaches 77.83\% average success on LIBERO, exceeding the strongest baseline in the main comparison by 9.50 percentage points.
The ablations support collecting expert execution inputs, retaining intermediate generation calls, and replaying merged features within the tested recipe.
Continued training from TCR reaches 97.75\% peak success, versus 94.00\% when training from the common base.
\label{end:maintext}

\subsection*{AI use statement}
Generative AI tools assisted with manuscript editing.
The authors are responsible for verifying the manuscript, code, citations, and supporting artifacts.

\bibliographystyle{plainnat}
\bibliography{references}

\begin{thebibliography}{52}
\providecommand{\natexlab}[1]{#1}
\providecommand{\url}[1]{\texttt{#1}}
\expandafter\ifx\csname urlstyle\endcsname\relax
  \providecommand{\doi}[1]{doi: #1}\else
  \providecommand{\doi}{doi: \begingroup \urlstyle{rm}\Url}\fi

\bibitem[Black et~al.(2025{\natexlab{a}})Black, Brown, Darpinian, Dhabalia, Driess, Esmail, Equi, Finn, Fusai, Galliker, Ghosh, Groom, Hausman, ichter, Jakubczak, Jones, Ke, LeBlanc, Levine, Li-Bell, Mothukuri, Nair, Pertsch, Ren, Shi, Smith, Springenberg, Stachowicz, Tanner, Vuong, Walke, Walling, Wang, Yu, and Zhilinsky]{pmlr-v305-black25a}
Kevin Black, Noah Brown, James Darpinian, Karan Dhabalia, Danny Driess, Adnan Esmail, Michael~Robert Equi, Chelsea Finn, Niccolo Fusai, Manuel~Y. Galliker, Dibya Ghosh, Lachy Groom, Karol Hausman, brian ichter, Szymon Jakubczak, Tim Jones, Liyiming Ke, Devin LeBlanc, Sergey Levine, Adrian Li-Bell, Mohith Mothukuri, Suraj Nair, Karl Pertsch, Allen~Z. Ren, Lucy~Xiaoyang Shi, Laura Smith, Jost~Tobias Springenberg, Kyle Stachowicz, James Tanner, Quan Vuong, Homer Walke, Anna Walling, Haohuan Wang, Lili Yu, and Ury Zhilinsky.
\newblock $\pi_{0.5}$: a vision-language-action model with open-world generalization.
\newblock In Joseph Lim, Shuran Song, and Hae-Won Park, editors, \emph{Proceedings of The 9th Conference on Robot Learning}, volume 305 of \emph{Proceedings of Machine Learning Research}, pages 17--40. PMLR, 27--30 Sep 2025{\natexlab{a}}.
\newblock URL \url{https://proceedings.mlr.press/v305/black25a.html}.

\bibitem[Black et~al.(2025{\natexlab{b}})Black, Brown, Driess, Esmail, Equi, Finn, Fusai, Groom, Hausman, Ichter, Jakubczak, Jones, Ke, Levine, Li-Bell, Mothukuri, Nair, Pertsch, Shi, Smith, Tanner, Vuong, Walling, Wang, and Zhilinsky]{BlackK-RSS-25}
Kevin Black, Noah Brown, Danny Driess, Adnan Esmail, Michael~Robert Equi, Chelsea Finn, Niccolo Fusai, Lachy Groom, Karol Hausman, Brian Ichter, Szymon Jakubczak, Tim Jones, Liyiming Ke, Sergey Levine, Adrian Li-Bell, Mohith Mothukuri, Suraj Nair, Karl Pertsch, Lucy~Xiaoyang Shi, Laura Smith, James Tanner, Quan Vuong, Anna Walling, Haohuan Wang, and Ury Zhilinsky.
\newblock {π₀: A Vision-Language-Action Flow Model for General Robot Control}.
\newblock In \emph{Proceedings of Robotics: Science and Systems}, LosAngeles, CA, USA, June 2025{\natexlab{b}}.
\newblock \doi{10.15607/RSS.2025.XXI.010}.

\bibitem[Chen et~al.(2025)Chen, Chen, Chen, Cai, Liu, Li, Liang, Lin, Ge, Gu, Deng, Guo, Nian, Xie, Chen, Su, Xu, Liu, Hu, ang Gao, Wang, Liang, Qin, Yang, Luo, and Mu]{chen2025robotwin20scalabledata}
Tianxing Chen, Zanxin Chen, Baijun Chen, Zijian Cai, Yibin Liu, Zixuan Li, Qiwei Liang, Xianliang Lin, Yiheng Ge, Zhenyu Gu, Weiliang Deng, Yubin Guo, Tian Nian, Xuanbing Xie, Qiangyu Chen, Kailun Su, Tianling Xu, Guodong Liu, Mengkang Hu, Huan ang Gao, Kaixuan Wang, Zhixuan Liang, Yusen Qin, Xiaokang Yang, Ping Luo, and Yao Mu.
\newblock Robotwin 2.0: A scalable data generator and benchmark with strong domain randomization for robust bimanual robotic manipulation, 2025.
\newblock URL \url{https://arxiv.org/abs/2506.18088}.

\bibitem[Cheng et~al.(2025)Cheng, Xiong, Wei, Zhu, and Yuan]{cheng2025whoever}
Runxi Cheng, Feng Xiong, Yongxian Wei, Wanyun Zhu, and Chun Yuan.
\newblock Whoever started the interference should end it: Guiding data-free model merging via task vectors.
\newblock In Aarti Singh, Maryam Fazel, Daniel Hsu, Simon Lacoste-Julien, Felix Berkenkamp, Tegan Maharaj, Kiri Wagstaff, and Jerry Zhu, editors, \emph{Proceedings of the 42nd International Conference on Machine Learning}, volume 267 of \emph{Proceedings of Machine Learning Research}, pages 10121--10143. PMLR, 13--19 Jul 2025.
\newblock URL \url{https://proceedings.mlr.press/v267/cheng25h.html}.

\bibitem[Chi et~al.(2023)Chi, Feng, Du, Xu, Cousineau, Burchfiel, and Song]{Chi-RSS-23}
Cheng Chi, Siyuan Feng, Yilun Du, Zhenjia Xu, Eric Cousineau, Benjamin~CM Burchfiel, and Shuran Song.
\newblock {Diffusion Policy: Visuomotor Policy Learning via Action Diffusion}.
\newblock In \emph{Proceedings of Robotics: Science and Systems}, Daegu, Republic of Korea, July 2023.
\newblock \doi{10.15607/RSS.2023.XIX.026}.

\bibitem[Davari and Belilovsky(2025)]{10.1007/978-3-031-73226-3_16}
MohammadReza Davari and Eugene Belilovsky.
\newblock Model breadcrumbs: Scaling multi-task model merging with sparse masks.
\newblock In Ale{\v{s}} Leonardis, Elisa Ricci, Stefan Roth, Olga Russakovsky, Torsten Sattler, and G{\"u}l Varol, editors, \emph{Computer Vision -- ECCV 2024}, pages 270--287, Cham, 2025. Springer Nature Switzerland.
\newblock ISBN 978-3-031-73226-3.

\bibitem[Deep et~al.(2024)Deep, Bhardwaj, and Poria]{deep2024dellamergingreducinginterferencemodel}
Pala~Tej Deep, Rishabh Bhardwaj, and Soujanya Poria.
\newblock Della-merging: Reducing interference in model merging through magnitude-based sampling, 2024.
\newblock URL \url{https://arxiv.org/abs/2406.11617}.

\bibitem[Dey et~al.(2025)Dey, Zaech, Nikolov, Van~Gool, and Paudel]{11128823}
Sombit Dey, Jan-Nico Zaech, Nikolay Nikolov, Luc Van~Gool, and Danda~Pani Paudel.
\newblock Revla: Reverting visual domain limitation of robotic foundation models.
\newblock In \emph{2025 IEEE International Conference on Robotics and Automation (ICRA)}, pages 8679--8686, 2025.
\newblock \doi{10.1109/ICRA55743.2025.11128823}.

\bibitem[Fu et~al.(2026)Fu, Zhang, Zhang, Wang, Huang, and Luo]{Fu_2026_CVPR}
Yuxia Fu, Zhizhen Zhang, Yuqi Zhang, Zijian Wang, Zi~Huang, and Yadan Luo.
\newblock Mergevla: Cross-skill model merging toward a generalist vision-language-action agent.
\newblock In \emph{Proceedings of the IEEE/CVF Conference on Computer Vision and Pattern Recognition (CVPR)}, pages 22335--22347, June 2026.

\bibitem[Gargiulo et~al.(2025)Gargiulo, Crisostomi, Bucarelli, Scardapane, Silvestri, and Rodolà]{11092448}
Antonio~Andrea Gargiulo, Donato Crisostomi, Maria~Sofia Bucarelli, Simone Scardapane, Fabrizio Silvestri, and Emanuele Rodolà.
\newblock Task singular vectors: Reducing task interference in model merging.
\newblock In \emph{2025 IEEE/CVF Conference on Computer Vision and Pattern Recognition (CVPR)}, pages 18695--18705, 2025.
\newblock \doi{10.1109/CVPR52734.2025.01742}.

\bibitem[Gu et~al.(2026)Gu, Cai, Wang, Wang, Wang, Wang, Huang, Lu, Wu, and Yang]{gu2026featcalfeaturecalibrationpostmerging}
Yanggan Gu, Shuo Cai, Zihao Wang, Wenjun Wang, Yuanyi Wang, Pengkai Wang, Sirui Huang, Su~Lu, Jianmin Wu, and Hongxia Yang.
\newblock Featcal: Feature calibration for post-merging models, 2026.
\newblock URL \url{https://arxiv.org/abs/2605.13030}.

\bibitem[Hu et~al.(2022)Hu, yelong shen, Wallis, Allen-Zhu, Li, Wang, Wang, and Chen]{hu2022lora}
Edward~J Hu, yelong shen, Phillip Wallis, Zeyuan Allen-Zhu, Yuanzhi Li, Shean Wang, Lu~Wang, and Weizhu Chen.
\newblock Lo{RA}: Low-rank adaptation of large language models.
\newblock In \emph{International Conference on Learning Representations}, 2022.
\newblock URL \url{https://openreview.net/forum?id=nZeVKeeFYf9}.

\bibitem[Ilharco et~al.(2023)Ilharco, Ribeiro, Wortsman, Schmidt, Hajishirzi, and Farhadi]{ilharco2023editing}
Gabriel Ilharco, Marco~Tulio Ribeiro, Mitchell Wortsman, Ludwig Schmidt, Hannaneh Hajishirzi, and Ali Farhadi.
\newblock Editing models with task arithmetic.
\newblock In \emph{The Eleventh International Conference on Learning Representations}, 2023.
\newblock URL \url{https://openreview.net/forum?id=6t0Kwf8-jrj}.

\bibitem[Jia et~al.(2026{\natexlab{a}})Jia, Jin, Zhang, Fan, Liang, Wang, Jin, Yu, Zhang, Song, Chen, and Yue]{jia2026bodymoveslearninganticipatory}
Haozhe Jia, Honglei Jin, Yuan Zhang, Youcheng Fan, Shaofeng Liang, Lei Wang, Shuxu Jin, Kuimou Yu, Zinuo Zhang, Jianfei Song, Wenshuo Chen, and Yutao Yue.
\newblock Before the body moves: Learning anticipatory joint intent for language-conditioned humanoid control, 2026{\natexlab{a}}.
\newblock URL \url{https://arxiv.org/abs/2605.14417}.

\bibitem[Jia et~al.(2026{\natexlab{b}})Jia, Song, Zhang, Jin, Fan, Chen, Zhang, and Yue]{jia2026echoedgecloudhumanoidorchestration}
Haozhe Jia, Jianfei Song, Yuan Zhang, Honglei Jin, Youcheng Fan, Wenshuo Chen, Wei Zhang, and Yutao Yue.
\newblock Echo: Edge-cloud humanoid orchestration for language-to-motion control, 2026{\natexlab{b}}.
\newblock URL \url{https://arxiv.org/abs/2603.16188}.

\bibitem[Jin et~al.(2023)Jin, Ren, Preotiuc-Pietro, and Cheng]{jin2023dataless}
Xisen Jin, Xiang Ren, Daniel Preotiuc-Pietro, and Pengxiang Cheng.
\newblock Dataless knowledge fusion by merging weights of language models.
\newblock In \emph{The Eleventh International Conference on Learning Representations}, 2023.
\newblock URL \url{https://openreview.net/forum?id=FCnohuR6AnM}.

\bibitem[Kim et~al.(2025{\natexlab{a}})Kim, Finn, and Liang]{KimM1-RSS-25}
Moo~Jin Kim, Chelsea Finn, and Percy Liang.
\newblock {Fine-Tuning Vision-Language-Action Models: Optimizing Speed and Success}.
\newblock In \emph{Proceedings of Robotics: Science and Systems}, LosAngeles, CA, USA, June 2025{\natexlab{a}}.
\newblock \doi{10.15607/RSS.2025.XXI.017}.

\bibitem[Kim et~al.(2025{\natexlab{b}})Kim, Pertsch, Karamcheti, Xiao, Balakrishna, Nair, Rafailov, Foster, Sanketi, Vuong, Kollar, Burchfiel, Tedrake, Sadigh, Levine, Liang, and Finn]{pmlr-v270-kim25c}
Moo~Jin Kim, Karl Pertsch, Siddharth Karamcheti, Ted Xiao, Ashwin Balakrishna, Suraj Nair, Rafael Rafailov, Ethan~P Foster, Pannag~R Sanketi, Quan Vuong, Thomas Kollar, Benjamin Burchfiel, Russ Tedrake, Dorsa Sadigh, Sergey Levine, Percy Liang, and Chelsea Finn.
\newblock Openvla: An open-source vision-language-action model.
\newblock In Pulkit Agrawal, Oliver Kroemer, and Wolfram Burgard, editors, \emph{Proceedings of The 8th Conference on Robot Learning}, volume 270 of \emph{Proceedings of Machine Learning Research}, pages 2679--2713. PMLR, 06--09 Nov 2025{\natexlab{b}}.
\newblock URL \url{https://proceedings.mlr.press/v270/kim25c.html}.

\bibitem[Lee et~al.(2026)Lee, Jeong, and Yoon]{Lee_2026_CVPR}
Wonyoung Lee, Wooseong Jeong, and Kuk-Jin Yoon.
\newblock Label-free cross-task lora merging with null-space compression.
\newblock In \emph{Proceedings of the IEEE/CVF Conference on Computer Vision and Pattern Recognition (CVPR)}, pages 847--859, June 2026.

\bibitem[Li et~al.(2026)Li, Qi, Tian, and Geng]{Li_2026_CVPR}
Longhua Li, Lei Qi, Qi~Tian, and Xin Geng.
\newblock Model merging in the essential subspace.
\newblock In \emph{Proceedings of the IEEE/CVF Conference on Computer Vision and Pattern Recognition (CVPR)}, pages 31145--31154, June 2026.

\bibitem[Lipman et~al.(2023)Lipman, Chen, Ben-Hamu, Nickel, and Le]{lipman2023flow}
Yaron Lipman, Ricky T.~Q. Chen, Heli Ben-Hamu, Maximilian Nickel, and Matthew Le.
\newblock Flow matching for generative modeling.
\newblock In \emph{The Eleventh International Conference on Learning Representations}, 2023.
\newblock URL \url{https://openreview.net/forum?id=PqvMRDCJT9t}.

\bibitem[Liu et~al.(2023)Liu, Zhu, Gao, Feng, Liu, Zhu, and Stone]{NEURIPS2023_8c3c6668}
Bo~Liu, Yifeng Zhu, Chongkai Gao, Yihao Feng, Qiang Liu, Yuke Zhu, and Peter Stone.
\newblock Libero: Benchmarking knowledge transfer for lifelong robot learning.
\newblock In A.~Oh, T.~Naumann, A.~Globerson, K.~Saenko, M.~Hardt, and S.~Levine, editors, \emph{Advances in Neural Information Processing Systems}, volume~36, pages 44776--44791. Curran Associates, Inc., 2023.
\newblock \doi{10.52202/075280-1939}.
\newblock URL \url{https://proceedings.neurips.cc/paper_files/paper/2023/file/8c3c666820ea055a77726d66fc7d447f-Paper-Datasets_and_Benchmarks.pdf}.

\bibitem[Marczak et~al.(2025)Marczak, Magistri, Cygert, Twardowski, Bagdanov, and van~de Weijer]{marczak2025no}
Daniel Marczak, Simone Magistri, Sebastian Cygert, Bart{\l}omiej Twardowski, Andrew~D. Bagdanov, and Joost van~de Weijer.
\newblock No task left behind: Isotropic model merging with common and task-specific subspaces.
\newblock In \emph{Forty-second International Conference on Machine Learning}, 2025.
\newblock URL \url{https://openreview.net/forum?id=RBZpAa27ls}.

\bibitem[Matena and Raffel(2022)]{NEURIPS2022_70c26937}
Michael~S Matena and Colin Raffel.
\newblock Merging models with fisher-weighted averaging.
\newblock In S.~Koyejo, S.~Mohamed, A.~Agarwal, D.~Belgrave, K.~Cho, and A.~Oh, editors, \emph{Advances in Neural Information Processing Systems}, volume~35, pages 17703--17716. Curran Associates, Inc., 2022.
\newblock \doi{10.52202/068431-1287}.
\newblock URL \url{https://proceedings.neurips.cc/paper_files/paper/2022/file/70c26937fbf3d4600b69a129031b66ec-Paper-Conference.pdf}.

\bibitem[Nguyen et~al.(2026)Nguyen, Huu-Tien, Suzuki, and Nguyen]{nguyen2026regmean}
The-Hai Nguyen, Dang Huu-Tien, Takeshi Suzuki, and Le-Minh Nguyen.
\newblock Regmean++: Enhancing effectiveness and generalization of regression mean for model merging.
\newblock \emph{Transactions on Machine Learning Research}, 2026.
\newblock ISSN 2835-8856.
\newblock URL \url{https://openreview.net/forum?id=H5lDsSCS9i}.

\bibitem[Ross et~al.(2011)Ross, Gordon, and Bagnell]{pmlr-v15-ross11a}
Stephane Ross, Geoffrey Gordon, and Drew Bagnell.
\newblock A reduction of imitation learning and structured prediction to no-regret online learning.
\newblock In Geoffrey Gordon, David Dunson, and Miroslav Dudík, editors, \emph{Proceedings of the Fourteenth International Conference on Artificial Intelligence and Statistics}, volume~15 of \emph{Proceedings of Machine Learning Research}, pages 627--635, Fort Lauderdale, FL, USA, 11--13 Apr 2011. PMLR.
\newblock URL \url{https://proceedings.mlr.press/v15/ross11a.html}.

\bibitem[Rusu et~al.(2016)Rusu, Colmenarejo, Gülçehre, Desjardins, Kirkpatrick, Pascanu, Mnih, Kavukcuoglu, and Hadsell]{rusu2016iclr-policy}
Andrei~A. Rusu, Sergio~Gomez Colmenarejo, Çaglar Gülçehre, Guillaume Desjardins, James Kirkpatrick, Razvan Pascanu, Volodymyr Mnih, Koray Kavukcuoglu, and Raia Hadsell.
\newblock {Policy Distillation}.
\newblock In \emph{International Conference on Learning Representations}, 2016.
\newblock URL \url{https://mlanthology.org/iclr/2016/rusu2016iclr-policy/}.

\bibitem[Ryum and Kang(2026)]{ryum2026ctmergingconsensusdirectionstasklevel}
Keumseo Ryum and Joonhyuk Kang.
\newblock Ct-merging: Consensus directions and task-level scaling for lora adapter merging, 2026.
\newblock URL \url{https://arxiv.org/abs/2607.20561}.

\bibitem[Shi et~al.(2026)Shi, Zhou, Wang, Miao, and Zhao]{shi2026pactpreservinganchoredcores}
Ningyuan Shi, Zhipeng Zhou, Hao Wang, Chunyan Miao, and Peilin Zhao.
\newblock Pact: Preserving anchored cores in task-vectors for model merging, 2026.
\newblock URL \url{https://arxiv.org/abs/2606.18627}.

\bibitem[Shukor et~al.(2025)Shukor, Aubakirova, Capuano, Kooijmans, Palma, Zouitine, Aractingi, Pascal, Russi, Marafioti, Alibert, Cord, Wolf, and Cadene]{shukor2025smolvlavisionlanguageactionmodelaffordable}
Mustafa Shukor, Dana Aubakirova, Francesco Capuano, Pepijn Kooijmans, Steven Palma, Adil Zouitine, Michel Aractingi, Caroline Pascal, Martino Russi, Andres Marafioti, Simon Alibert, Matthieu Cord, Thomas Wolf, and Remi Cadene.
\newblock Smolvla: A vision-language-action model for affordable and efficient robotics, 2025.
\newblock URL \url{https://arxiv.org/abs/2506.01844}.

\bibitem[Si et~al.(2026)Si, Lu, Yang, Ye, Wu, and Jiang]{si2026vlaprocrosstaskproceduralmemory}
Shengyu Si, Yuanzhuo Lu, Ruimeng Yang, Ziyi Ye, Zuxuan Wu, and Yu-Gang Jiang.
\newblock Vla-pro: Cross-task procedural memory transfer for vision-language-action models, 2026.
\newblock URL \url{https://arxiv.org/abs/2605.29562}.

\bibitem[Skorobogat et~al.(2025)Skorobogat, Roth, and Georgescu]{skorobogat2025subspaceboostedmodelmerging}
Ronald Skorobogat, Karsten Roth, and Mariana-Iuliana Georgescu.
\newblock Subspace-boosted model merging, 2025.
\newblock URL \url{https://arxiv.org/abs/2506.16506}.

\bibitem[Song et~al.(2026)Song, Zhao, Li, Zhou, Wang, Lyu, Ding, Wang, Wang, and Li]{song2026capvectorlearningtransferablecapability}
Wenxuan Song, Han Zhao, Fuhao Li, Ziyang Zhou, Xi~Wang, Jing Lyu, Pengxiang Ding, Yan Wang, Donglin Wang, and Haoang Li.
\newblock Capvector: Learning transferable capability vectors in parametric space for vision-language-action models, 2026.
\newblock URL \url{https://arxiv.org/abs/2605.10903}.

\bibitem[Stoica et~al.(2025)Stoica, Ramesh, Ecsedi, Choshen, and Hoffman]{stoica2025model}
George Stoica, Pratik Ramesh, Boglarka Ecsedi, Leshem Choshen, and Judy Hoffman.
\newblock Model merging with svd to tie the knots.
\newblock In \emph{International Conference on Learning Representations}, volume 2025, pages 4501--4519, 2025.

\bibitem[Sun et~al.(2025)Sun, Li, Wang, Liu, Geng, and Li]{NEURIPS2025_fbb4f226}
Wenju Sun, Qingyong Li, Wen Wang, Yang Liu, Yangliao Geng, and Boyang Li.
\newblock Towards minimizing feature drift in model merging: Layer-wise task vector fusion for adaptive knowledge integration.
\newblock In D.~Belgrave, C.~Zhang, H.~Lin, R.~Pascanu, P.~Koniusz, M.~Ghassemi, and N.~Chen, editors, \emph{Advances in Neural Information Processing Systems}, volume 38, Main Conference, pages 172635--172663. Curran Associates, Inc., 2025.
\newblock \doi{10.52202/085713-5745}.
\newblock URL \url{https://proceedings.neurips.cc/paper_files/paper/2025/file/fbb4f2269267de23e0240e1c01242981-Paper-Conference.pdf}.

\bibitem[Tian et~al.(2024)Tian, Lai, Wu, Shuai, Ge, and Yue]{bowen2024taskvectorsselectivetask}
Bowen Tian, Songning Lai, Jiemin Wu, Zhihao Shuai, Shiming Ge, and Yutao Yue.
\newblock Beyond task vectors: Selective task arithmetic based on importance metrics, 2024.
\newblock URL \url{https://arxiv.org/abs/2411.16139}.

\bibitem[Tian et~al.(2025)Tian, Chen, Li, Lai, Wu, and Yue]{10.1145/3746027.3755441}
Bowen Tian, Wenshuo Chen, Zexi Li, Songning Lai, Jiemin Wu, and Yutao Yue.
\newblock Text2weight: Bridging natural language and neural network weight spaces.
\newblock In \emph{Proceedings of the 33rd ACM International Conference on Multimedia}, MM '25, pages 10152--10160, New York, NY, USA, 2025. Association for Computing Machinery.
\newblock ISBN 9798400720352.
\newblock \doi{10.1145/3746027.3755441}.
\newblock URL \url{https://doi.org/10.1145/3746027.3755441}.

\bibitem[Vaina et~al.(2026)Vaina, Chumachenko, and Ryabinin]{locicerovaina2026mashup}
Sofia Maria Lo~Cicero Vaina, Artem Chumachenko, and Max Ryabinin.
\newblock Mashup learning: Faster finetuning by remixing past checkpoints.
\newblock \emph{arXiv preprint arXiv:2603.10156}, 2026.

\bibitem[Wang et~al.(2024)Wang, Zhang, Zhou, Simchowitz, and Tedrake]{ICLR2024_e28b3369}
Lirui Wang, Kaiqing Zhang, Allan Zhou, Max Simchowitz, and Russ Tedrake.
\newblock Robot fleet learning via policy merging.
\newblock In B.~Kim, Y.~Yue, S.~Chaudhuri, K.~Fragkiadaki, M.~Khan, and Y.~Sun, editors, \emph{International Conference on Learning Representations}, volume 2024, pages 51737--51771, 2024.
\newblock URL \url{https://proceedings.iclr.cc/paper_files/paper/2024/file/e28b3369186459f57c94a9ec9137fac9-Paper-Conference.pdf}.

\bibitem[Wang et~al.(2026)Wang, Hou, Liang, Tian, Zhang, Chen, and Yue]{wang-etal-2026-able}
Zirui Wang, Yusen Hou, Shaofeng Liang, Bowen Tian, Yanlin Zhang, Wenshuo Chen, and Yutao Yue.
\newblock {ABLE}: Representing and mapping {LLM}s via attribution-based large-model embedding.
\newblock In Hung-Yi Lee, Sunipa Dev, Junyi~Jessy Li, and Desmond Elliott, editors, \emph{Proceedings of the 2026 Conference on Empirical Methods in Natural Language Processing}, pages 46513--46536, Budapest, Hungary, October 2026. Association for Computational Linguistics.
\newblock ISBN 979-8-89176-459-0.
\newblock \doi{10.18653/2026.emnlp-main.2040}.
\newblock URL \url{https://aclanthology.org/2026.emnlp-main.2040/}.

\bibitem[Wei et~al.(2026)Wei, Cheng, Zhang, Shen, Yuan, Cui, and Tao]{wei2026closedformspectralregularizationmultitask}
Yongxian Wei, Runxi Cheng, Xingxuan Zhang, Li~Shen, Chun Yuan, Peng Cui, and Dacheng Tao.
\newblock Closed-form spectral regularization for multi-task model merging, 2026.
\newblock URL \url{https://arxiv.org/abs/2606.07289}.

\bibitem[Wortsman et~al.(2022)Wortsman, Ilharco, Gadre, Roelofs, Gontijo-Lopes, Morcos, Namkoong, Farhadi, Carmon, Kornblith, and Schmidt]{pmlr-v162-wortsman22a}
Mitchell Wortsman, Gabriel Ilharco, Samir~Ya Gadre, Rebecca Roelofs, Raphael Gontijo-Lopes, Ari~S Morcos, Hongseok Namkoong, Ali Farhadi, Yair Carmon, Simon Kornblith, and Ludwig Schmidt.
\newblock Model soups: averaging weights of multiple fine-tuned models improves accuracy without increasing inference time.
\newblock In Kamalika Chaudhuri, Stefanie Jegelka, Le~Song, Csaba Szepesvari, Gang Niu, and Sivan Sabato, editors, \emph{Proceedings of the 39th International Conference on Machine Learning}, volume 162 of \emph{Proceedings of Machine Learning Research}, pages 23965--23998. PMLR, 17--23 Jul 2022.
\newblock URL \url{https://proceedings.mlr.press/v162/wortsman22a.html}.

\bibitem[Yadav et~al.(2023)Yadav, Tam, Choshen, Raffel, and Bansal]{yadav2023tiesmerging}
Prateek Yadav, Derek Tam, Leshem Choshen, Colin~A Raffel, and Mohit Bansal.
\newblock Ties-merging: Resolving interference when merging models.
\newblock \emph{Advances in neural information processing systems}, 36:\penalty0 7093--7115, 2023.

\bibitem[Yadav et~al.(2026)Yadav, Zhou, Wagenmaker, Pertsch, and Levine]{yadav2026robust}
Yajat Yadav, Zhiyuan Zhou, Andrew Wagenmaker, Karl Pertsch, and Sergey Levine.
\newblock Robust fine-tuning of vision-language-action robot policies via parameter merging.
\newblock In \emph{International Conference on Learning Representations}, volume 2026, pages 56385--56417, 2026.

\bibitem[Yang et~al.(2024)Yang, Wang, Shen, Liu, Guo, Wang, and Tao]{yang2024adamerging}
Enneng Yang, Zhenyi Wang, Li~Shen, Shiwei Liu, Guibing Guo, Xingwei Wang, and Dacheng Tao.
\newblock Adamerging: Adaptive model merging for multi-task learning.
\newblock In \emph{International Conference on Learning Representations}, volume 2024, pages 22743--22763, 2024.

\bibitem[Yao et~al.(2025)Yao, LIU, Liu, Li, LIU, Han, Guo, Wu, and Song]{NEURIPS2025_b395e7bd}
Yuxuan Yao, Shuqi LIU, Zehua Liu, Qintong Li, Mingyang LIU, Xiongwei Han, Zhijiang Guo, Han Wu, and Linqi Song.
\newblock Activation-guided consensus merging for large language models.
\newblock In D.~Belgrave, C.~Zhang, H.~Lin, R.~Pascanu, P.~Koniusz, M.~Ghassemi, and N.~Chen, editors, \emph{Advances in Neural Information Processing Systems}, volume 38, Main Conference, pages 124041--124069. Curran Associates, Inc., 2025.
\newblock \doi{10.52202/085713-4133}.
\newblock URL \url{https://proceedings.neurips.cc/paper_files/paper/2025/file/b395e7bd5eaf3e980386debf2e9d7f11-Paper-Conference.pdf}.

\bibitem[Yu et~al.(2024)Yu, Yu, Yu, Huang, and Li]{yu2024language}
Le~Yu, Bowen Yu, Haiyang Yu, Fei Huang, and Yongbin Li.
\newblock Language models are super mario: absorbing abilities from homologous models as a free lunch.
\newblock In \emph{Proceedings of the 41st International Conference on Machine Learning}, ICML'24. JMLR.org, 2024.

\bibitem[Zhang et~al.(2026)Zhang, Zhou, Luo, Di, Zhang, and Wei]{zhang2026dcmergeimprovingmodelmerging}
Han-Chen Zhang, Zi-Hao Zhou, Mao-Lin Luo, Shimin Di, Min-Ling Zhang, and Tong Wei.
\newblock Dc-merge: Improving model merging with directional consistency, 2026.
\newblock URL \url{https://arxiv.org/abs/2603.06242}.

\bibitem[Zhang and Zhou(2025)]{zhang-zhou-2025-unraveling}
Haobo Zhang and Jiayu Zhou.
\newblock Unraveling {L}o{RA} interference: Orthogonal subspaces for robust model merging.
\newblock In Wanxiang Che, Joyce Nabende, Ekaterina Shutova, and Mohammad~Taher Pilehvar, editors, \emph{Proceedings of the 63rd Annual Meeting of the Association for Computational Linguistics (Volume 1: Long Papers)}, pages 26459--26472, Vienna, Austria, July 2025. Association for Computational Linguistics.
\newblock ISBN 979-8-89176-251-0.
\newblock \doi{10.18653/v1/2025.acl-long.1284}.
\newblock URL \url{https://aclanthology.org/2025.acl-long.1284/}.

\bibitem[Zhao et~al.(2023)Zhao, Kumar, Levine, and Finn]{Zhao-RSS-23}
Tony~Z. Zhao, Vikash Kumar, Sergey Levine, and Chelsea Finn.
\newblock {Learning Fine-Grained Bimanual Manipulation with Low-Cost Hardware}.
\newblock In \emph{Proceedings of Robotics: Science and Systems}, Daegu, Republic of Korea, July 2023.
\newblock \doi{10.15607/RSS.2023.XIX.016}.

\bibitem[Zheng et~al.(2025)Zheng, Wang, Huang, Wang, Chen, Fan, Hu, and Ye]{zheng2025decoupleorthogonalizedatafreeframework}
Shenghe Zheng, Hongzhi Wang, Chenyu Huang, Xiaohui Wang, Tao Chen, Jiayuan Fan, Shuyue Hu, and Peng Ye.
\newblock Decouple and orthogonalize: A data-free framework for lora merging, 2025.
\newblock URL \url{https://arxiv.org/abs/2505.15875}.

\bibitem[Zhou et~al.(2026)Zhou, Xu, Tie, Chen, Zhang, Chu, Zhou, and Sun]{zhou2026liberoprorobustfairevaluation}
Xueyang Zhou, Yangming Xu, Guiyao Tie, Yongchao Chen, Guowen Zhang, Duanfeng Chu, Pan Zhou, and Lichao Sun.
\newblock Libero-pro: Towards robust and fair evaluation of vision-language-action models beyond memorization, 2026.
\newblock URL \url{https://arxiv.org/abs/2510.03827}.

\end{thebibliography}

\appendix
\section{Method and Implementation Details}
\label{app:method}

Throughout, $i,r,f,b,j$ index experts, policy requests, generation calls, blocks, and linear modules; $m,t,e,k,g$ index methods, tasks, evaluation episodes, repeats, and task groups. A cache request index retains its episode identity. Module indices are suppressed in local regression formulas.

\subsection{Assumptions and Scope}
Experts should have homologous computation graphs and compatible input, action, tokenizer, and normalization conventions.
Each calibration state is associated offline with the expert whose task group generated it; this association is not an inference-time task ID and is absent from the deployed checkpoint.
Finally, the calibration set should cover the activation directions on which expert behavior differs: Eq.~\eqref{eq:tcr_solve} deliberately retains the prior in unsupported directions and therefore cannot infer a missing skill from parameter geometry alone.
With $N_{\mathrm{rows}}=\sum_in_i$, the sample-space solve uses $O(N_{\mathrm{rows}}^2)$ kernel storage and $O(N_{\mathrm{rows}}^3)$ factorization time per module, in addition to replay and weight reconstruction.

\paragraph{Checkpoint materialization.}
\label{app:materialization}
TCR takes effective expert weights as input, independently of their fine-tuning parameterization.
For an expert trained with LoRA, we materialize $W_i=W_0+\gamma_i B_iA_i$ before fusion, where $\gamma_i$ is its adapter scale~\citep{hu2022lora}.
Directly trained weights and biases enter the same dense checkpoint without this conversion.
Materialization preserves the expert's function and is an implementation step, not a LoRA-specific fusion rule or a separate novelty claim.
The merged correction is solved in weight space rather than constrained to the original adapter factors.
Every adapted tensor is accounted for by either a regression solve or a declared fallback; full adapted-parameter coverage does not mean that every base-model tensor is changed.

\subsection{The Relative-Risk Identity}
The inverse-prior-error masses defined in Eq.~\eqref{eq:expert_mass} and used in Eq.~\eqref{eq:tcr_objective} have an exact interpretation.
Let $E_i(C)=\|X_iC^\top-X_i(W_i-P)^\top\|_F^2$, equivalently the output-matching residual in Eq.~\eqref{eq:tcr_objective}; then $E_i(0)=n_i d_{\mathrm{out}}e_i^{\mathrm{prior}}$.
When the floor is inactive, $\omega_i^{\mathrm{rel}}=(1/e_i^{\mathrm{prior}})/Z_{\mathrm{rel}}$, where $Z_{\mathrm{rel}}=\sum_{i'=1}^M1/e_{i'}^{\mathrm{prior}}$, and
\begin{equation}
 \sum_i\frac{\omega_i^{\mathrm{rel}}}{n_i}E_i(C)
 =\sum_i\frac{E_i(C)}{Z_{\mathrm{rel}}\,n_i e_i^{\mathrm{prior}}}
 =\frac{d_{\mathrm{out}}}{Z_{\mathrm{rel}}}\sum_i\frac{E_i(C)}{E_i(0)} .
 \label{eq:tcr_relative_risk}
\end{equation}
The objective thus minimizes a sum of \emph{fractional} errors relative to the common prior, rather than raw residual energy, which one high-energy expert could otherwise dominate.
The floor clips this interpretation only for numerically negligible prior errors.

\subsection{Numerical and Implementation Choices}
When there are fewer input rows than feature dimensions, the regression can be solved using a smaller matrix.
Stack $\widetilde X_i=\sqrt{\omega_i/n_i}X_i$ and
$\widetilde Y_i=\sqrt{\omega_i/n_i}X_i(W_i-P)^\top$ by rows; then
\begin{equation}
 C^\star=\widetilde Y^\top
 (\widetilde X\widetilde X^\top+\lambda I)^{-1}\widetilde X.
 \label{eq:tcr_dual}
\end{equation}
This is equivalent to Eq.~\eqref{eq:tcr_solve}; it changes how the system is solved, not its objective.

The first pass sets $\lambda=0.05\max(\|\widetilde X\|_F^2/d_{\mathrm{in}},10^{-12})$.
The second pass reuses the recorded numerical ridge value for each module from its first-pass reference rather than recomputing it from the second cache.
For a biased module we augment $X$ with a constant column and append the bias vector $\mathbf b_{\mathrm{bias}}$ to $W$, applying the same ridge penalty to the augmented correction; $d_{\mathrm{in}}$ then denotes the augmented width.
A radial safeguard limits extrapolation from finite calibration data: with $R_{\mathrm{cap}}=3\max(M^{-1}\sum_i\|W_i-P\|_F,10^{-12})$,
\begin{equation}
 \rho=\min\!\left(1,\frac{R_{\mathrm{cap}}}{\max(\|C^\star\|_F,10^{-12})}\right),
 \qquad W^{\mathrm M}=P+\rho\,C^\star .
 \label{eq:tcr_safeguard}
\end{equation}
This is a numerical safeguard applied after ridge regression, not the exact solution of a trust-constrained problem.
Whether it activates is checked per module and recorded in each checkpoint's construction manifest.
Equation~\eqref{eq:tcr_solve} follows from the normal equation for $Z=C^\top$ together with the push-through identity $(\widetilde X^\top\widetilde X+\lambda I)^{-1}\widetilde X^\top=\widetilde X^\top(\widetilde X\widetilde X^\top+\lambda I)^{-1}$, after transposing; positive ridge makes the system positive definite even when sampled activation rows are linearly dependent.
Retaining an immutable prior within a pass prevents the ridge center from depending on incidental module-update order.

\paragraph{Sampling and weighting.}
\label{app:calibration_recipes}
The main \pizerofive{} recipe uses two expert caches with different row-allocation rules.
Cache A supplies the first pass with 4,441,200 rows, using a cap of ten rows per stored call and view in its replay path.
Cache B uses ten tasks, one episode per task, five requests per episode, and three of ten flow steps, giving 150 cached call states per expert.
Its second-pass replay shares a cap of sixteen rows per request and module across selected stages, for 1,331,600 rows in total.
The cumulative budget is 5,772,800 rows; a cap value alone is not a comparable budget across these replay paths.
All retained flow states belonging to a sampled request are kept together, preserving the association between a robot observation and its denoising path.
Because each state contributes sampled token rows, equal state counts need not imply equal regression budgets when sequence lengths or module shapes differ, so we record the realized row count for every expert and module.
Within each expert, sample-source weights scale rows before regression. With unit source weights, sampled rows contribute equally, while the separate $\omega_i/n_i$ factor gives each expert its prescribed total objective mass.
The main comparison holds the expert bank and evaluation protocol fixed while retaining each method's calibration recipe and reporting its budget.
Component-removal studies report the rows lost with removed inputs.

\subsection{\pizerofive{} Replay Structure}
\label{app:replay_structure}
Table~\ref{tab:architecture} gives the per-stage breakdown of the 422 modified tensors.
Language replay refreshes the prefix keys and values consumed by the action expert; the two time projections are solved sequentially through their intervening activation; and the action output projection is solved once the final action hidden state is available.
Replaying a block is validated against the native forward pass under matched weights and inputs.
Comparing representations generated by \emph{different} policies is a distribution diagnostic rather than an equivalence test, and we do not use it as one.
Each artifact records expert and prior identities, replay manifests, per-module row counts, solver configuration, and checkpoint hashes.

\begin{table}[h]
\centering
\caption{\pizerofive{} replay structure and adapted linear modules. The output interface is solved after the action blocks. Identical common-base parameters outside this scope are retained.}
\label{tab:architecture}
\resulttable
\begin{tabular}{lrrl}
\toprule
Stage & Blocks & Linear & Updated boundary\\
\midrule
Vision & 27 & 162 & Visual tokens\\
Language & 18 & 126 & Prefix states, K/V\\
Action/time input & -- & 3 & Action tokens, time\\
Action expert & 18 & 126 & Action hidden states\\
Action output & -- & 1 & Predicted velocity\\
\midrule
Total & 63 & 418 & 422 tensors with bias\\
\bottomrule
\end{tabular}
\end{table}

\subsection{Fusion Procedure}
Preparation materializes every LoRA update into its dense expert weight, loads directly adapted interfaces, and verifies architecture, tokenizer, normalizers, and action conventions.
The first pass fixes the dense expert mean as its prior; the second fixes the first-pass checkpoint as its prior.
Each pass visits all adapted blocks in computation order using its designated expert cache and weighting rule.
For each block, the boundary is generated with the current merged prefix; each candidate block is run on its associated calibration states; linear inputs are captured; the prior-centered regression is solved; all solutions are written; and the completed block is replayed once to produce the next boundary.
Deployment checks tensor scope, finite values, configuration, and artifact hashes, then exports the merged weights with the original policy preprocessing.
We preserve normalizers, tokenizer settings, action-chunk length, and task text across every evaluated method.
The final static model needs no calibration source or task ID, although the natural-language instruction remains an ordinary policy input.

\subsection{Implementation Scope}
\label{app:backbone_contract}
The evaluated implementation uses \pizerofive{} observations, proprioception, instructions, noisy action chunks, and flow times. Its replay covers the vision and language prefixes, action conditioning, and action blocks. We verify replay against the native forward pass at matched weights and inputs before merging.

Extending TCR to another architecture would require a separate recording and replay adapter for the inputs and computation used at inference. In particular, a WAM evaluation would have to specify whether video generation is part of the executed policy and validate the corresponding replay. We do not report such a comparison here.

\FloatBarrier
\section{Experimental Protocols}
\label{app:protocol}

This appendix details the evaluation and comparison protocols.

\subsection{Main LIBERO Evaluation Protocol}
The unified comparison uses four experts adapted for 10{,}000 steps from one local \pizerofive{} base.
Every evaluated method shares the normalizers, tokenizer, ten-step flow sampler, ten executed actions per policy request, action-chunk length, and task text of the source experts, and is scored by the same evaluator on ten episodes for each of ten tasks per suite, i.e.\ 400 episodes per checkpoint.
Calibration inputs and development states are distinct from the formal procedural reset selections.
The procedural reset bank was fixed for the reported comparisons but had been inspected during development; reset identities and random-number settings are recorded in the artifacts.
Each method is evaluated three times, with 400 episodes per run; we report mean and sample standard deviation, using paired tasks and initial conditions across methods.
Registered random seeds do not guarantee identical generation-noise streams throughout diverging rollouts.
The primary main-table contrasts compare \methodshort{} with Model Soups and the strongest external baseline selected on development data.
The uniform first-pass weighting control is a separate mechanism contrast for error weighting.
The comparisons are conditional on the evaluated expert bank; they do not estimate variation across independently trained expert sets.
Calibration resampling, repeated policy evaluation, and independent expert training are recorded as separate experimental factors.
Let $\mathcal T$ be the task set, $E_t$ the episode count for task $t$, and $y_{t,e}(\pi)\in\{0,1\}$ the benchmark-defined success of policy $\pi$ in episode $e$.
We report task-macro success $\operatorname{Succ}(\pi)=|\mathcal T|^{-1}\sum_{t\in\mathcal T}E_t^{-1}\sum_{e=1}^{E_t}y_{t,e}(\pi)$, expressed as a percentage, together with per-suite success.
The method configuration is fixed before evaluation, and no repeat is selected by its success rate.

\subsection{Baseline Configurations}
\label{app:baseline_configs}
This section documents the baseline configurations; method citations appear in the main text.
The main table's RegMean++ uses sequential merged-block replay and the original off-diagonal Gram scaling with fixed $\alpha=0.3$, plus explicitly added spectral smoothing.
Calibration uses 200 static demonstration observations with initial Gaussian action noise at flow time $\tau_{\mathrm{flow}}=1$, without demonstration action labels or later generation states.
For a Gram eigenvalue $\sigma$, the inverse filter is $\sigma/(\sigma^2+\tau_{\mathrm{spec}}^2)$ on the correction around the arithmetic expert mean, with $\tau_{\mathrm{spec}}=d_{\mathrm{in}}\,\epsilon_{\mathrm{FP32}}\widehat\sigma_{\max}$ fixed by input width and estimated spectral scale before evaluation.
This is a regularized extension, not an algebraically equivalent implementation of the unmodified solver; it uses neither an additive ridge penalty nor a correction-norm cap, and biases use the expert mean.
RegMean++ yields 698 successes in 1,200 episodes ($58.17\pm1.04\%$).
The ordinary RegMean row uses each expert's own prefix and the same static observation pool, scope, and disclosed spectral smoothing rule as RegMean++.
It fixes $\alpha=0.3$ to align their Gram scaling; the remaining prefix propagation distinguishes the two methods.
RegMean yields 644 successes in 1,200 episodes ($53.67\pm0.63\%$).
The WUDI entry retains the original WUDI objective on 414 materialized task-vector matrices, followed by per-tensor RMS rescaling; the eight directly adapted action/time heads use dense averaging.
It applies 300 Adam updates per optimized matrix: 409 nonzero matrices are optimized, while five zero-delta matrices and the eight heads receive no gradient updates. These are construction updates, not end-to-end policy-training steps.
Model Soups uses uniform averaging~\citep{pmlr-v162-wortsman22a}: $\bar\Theta=M^{-1}\sum_{i=1}^{M}\Theta_i$, where each $\Theta_i$ is a materialized expert checkpoint.
It uses neither greedy soup selection nor averaging of separate LoRA factors.

Tables use the standard method names; the VLA adaptations and numerical settings are specified in this section.
Demonstration observations can supply native generator inputs without using demonstration actions as regression labels.

Method-specific construction details for real-robot evaluation, cross-benchmark comparisons, scaling, and further-training initializations are given in their respective protocols.
The main method comparison fixes the expert bank and evaluation settings, retaining and documenting each method's construction recipe.
Matched-row controls are reported separately.
The hardware comparison shares physical expert traces, as specified in Appendix~\ref{app:real_robot}.

KnOTS-TIES applies KnOTS alignment before TIES~\citep{stoica2025model,yadav2023tiesmerging}.
It operates on the materialized LoRA updates; directly adapted interface tensors use a predeclared fallback whose parameter coverage is reported.

\paragraph{FeatCal.}
The FeatCal row uses a VLA adaptation of the published feature-calibration objective and forward-order procedure, initialized from Model Soups~\citep{gu2026featcalfeaturecalibrationpostmerging}.
For the displayed result, FeatCal starts from the average of the 10,000-step experts; the base model remains available for its anchor construction.
It uses 200 demonstration observations with three independent Gaussian action latents per observation, all at flow time $\tau_{\mathrm{flow}}=1$, giving 4,441,200 actual regression rows.
Demonstration action values are not used; the three noise replicas are not three distinct generation stages.
TCR uses the same expert bank but two expert execution caches and 5,772,800 cumulative rows; its second-pass prior is its own first-pass checkpoint.
FeatCal pairs features from experts and the model being calibrated using the same records and row indices within its own solve.
It retains target interpolation, the merged/base anchor, inverse feature-statistic scale weights, and its own regularization rule.
These are not replaced by TCR's expert targets or relative-error weights.
The reported configuration fixes interpolation $\alpha_{\mathrm{FC}}=0.3$, anchor coefficient $\rho_{\mathrm{FC}}=2$, and regularization $\lambda_{\mathrm{FC}}=0.05$ before formal evaluation, without retuning on these outcomes.
The port calibrates 418 linear weights across 47 forward stages using FP64 solves; biases, LayerNorm parameters, and other uncalibrated tensors retain their Model Soups values.
This adapted parameter scope and initial-noise input convention are not an official released \pizerofive{} configuration.
Construction accounting includes the initial average, feature collection, calibration, and validation search; sharing cached inputs does not erase their collection cost.
This comparison evaluates complete recipes with a shared expert bank and different calibration budgets.
Ablations with a fixed solver are still required to isolate execution coverage.

\paragraph{Ablation implementation and attribution.}
TCR ablations share a fixed expert bank and prior-centered solver; the intervention determines which inputs or calibrated modules may change.
We solve ridge regression on sampled rows with bias augmentation and a declared regularization schedule.
A post-solve safeguard limits the weight correction.
The replay adapter follows multimodal dependencies, including conditioning keys and values and time projections (Appendix~\ref{app:method}).
These choices define our implementation.
TCR builds on regression and depth replay by sampling across interaction requests and native generation calls, using relative-error masses in the first pass and uniform masses in the second.

\paragraph{Configurations and grids.}
We select one global configuration per tunable method on the fixed development bank.
Task Arithmetic and TIES screen 21 and 69 candidates, respectively, on 40 development episodes per candidate.
WUDI screens two candidates on 40 episodes each; KnOTS-TIES validates one recipe on 400, RegMean and RegMean++ use 40 each, and FeatCal runs one fixed 40-episode development gate.
For KnOTS-TIES, RegMean, and RegMean++, the evaluated scale, rank, pruning, and regularization settings are recorded where applicable.
Each search grid and interface fallback is fixed before final evaluation.
The methods have different hyperparameters and search sizes.
We therefore disclose available validation budgets and fixed-default results; selection budgets are not equal.
We use descriptive names for local finite-row, prior-centered ports and do not label a control as a distinct RegMean++ baseline when its calibration set, replay, weighting, and solver are identical to \methodshort{}.

\paragraph{Relation to MergeVLA.}
The released MergeVLA system trains Qwen2.5-0.5B experts with sparse task masks and cross-attention-only action blocks, then selects a stored task mask and action head with a test-time router~\citep{Fu_2026_CVPR}.
Its published Task Arithmetic, TIES, TSV, KnOTS, WUDI, and EMR numbers therefore measure merging experts constructed for that system, not post-hoc fusion of our fixed \pizerofive{} experts.
Its expert construction could complement TCR, but empirical compatibility requires a separate audit of masks, heads, and native inference.
We do not treat its published system scores as algorithm comparisons on our expert bank.

\subsection{Shared Protocol for Calibration Controls}
\label{app:ablation_protocol}
The component study in Table~\ref{tab:mechanism_ablation} and the supplementary studies ask different questions.
The former replaces observation source, generation inputs, or the prefix used to compute features.
The cache, budget, and recipe studies in Tables~\ref{tab:cache_sequence_controls}--\ref{tab:full_recipe_controls} explain choices within the final two-pass construction.
All use the same four 10k-step experts, Soup initialization, native inference settings, and full adapted scope unless scope is the named intervention.

\paragraph{References and repetition units.}
The three component contrasts use the same expert bank, calibration pools, and 400-episode repeat selections as the displayed Full recipe. Each configuration has three independently constructed models.
The full recipe uses 4,441,200 first-pass and 1,331,600 second-pass rows. Removing intermediate calls reduces the variant's first-pass rows to 1,480,400; the remaining per-call sampling rule is unchanged.

For the supplementary cache, budget, and recipe matrices, A$\to$B is one shared reference built at the main method's budgets.
The supplementary A/B/C caches are newly collected execution episodes, not the original caches used to construct the three main-table TCR models; the expert weights and collection protocol are unchanged.
The A parent and A$\to$B child are reused wherever specified, instead of fitting or selecting a new reference for each row.
In particular, A$\to$A, A$\to$B, and A$\to$C branch from the identical A checkpoint, so their second-pass cache comparisons do not change the first-pass model.
These supplementary comparisons use one construction per configuration and a fixed evaluation seed on the same 400 episodes, reporting success without a cross-seed standard deviation; their references are not the main-table three-repeat mean.
Calibration episodes, held-out requests, and evaluation resets are separated by their recorded state identities.

\paragraph{Inputs and numerical controls.}
Caches have the same task allocation and request-sampling rule; fresh caches use different episodes, not only different generation noise.
Requests are selected without conditioning on success.
The demonstration replacement changes observations, proprioception, and instructions, but still uses the frozen experts' native generation.
It does not substitute demonstration actions for intermediate states or regression targets.
Generation and prefix controls share the raw requests with Full.
An early-request control retains the request count; episodes with insufficient distinct requests are not padded with duplicates.

Every variant records per-module and per-expert row counts.
The named intervention determines whether rows, requests, or scope can change; other quotas remain fixed.
Matched-budget studies do not duplicate rows to meet a quota.
The reference's numerical ridge map is reused for each paired intervention rather than recomputed from a changed Gram scale.
The first-pass relative-error rule is unchanged except in the uniform-weight control; its realized masses can nevertheless change with inputs.
Second and later passes use uniform masses and the designated parent's weights as their fixed prior.
Precision, module order, bias treatment, and the correction safeguard are held fixed.
Changes to scope reduce total work and are not interpreted as equal-compute comparisons.

\paragraph{Measurements.}
Closed-loop evaluation uses the same tasks and reset selection within each contrast.
Matching seeds does not imply identical generation-noise streams after policies visit different states.
We report per-suite and aggregate success, paired episode wins and losses, and task-cluster uncertainty where available.
Action metrics use the common held-out bank defined below, not training residuals.
Construction records include actual rows and, where available, collection and solving time and peak memory.
No cache or configuration is selected by its final evaluation score.

\paragraph{Held-out action agreement.}
\label{app:action_fidelity}
These measurements reuse the checkpoints from each comparison; they are not an additional model-selection experiment.
Every checkpoint is measured on the same held-out request bank, and construction-level scores are aggregated only within their declared comparison.
These request-level measurements are separate from the 1,200 closed-loop episodes and do not require rerunning Full's success evaluations.
The common bank contains two expert episodes per task and five requests at fixed trajectory-time quantiles per episode, giving 400 requests over 40 tasks.
Initial-state content hashes are checked against the calibration and success-evaluation banks; selection does not condition on success or model disagreement.
Episode, request, input, and actual initial-noise identities are retained for each measurement.
Each merged policy and the corresponding task expert receives the same request and exact initial generation noise, but generates its own intermediate states using the native sampler and execution horizon.
We compare only the action slots that would actually be executed, not unused future predictions, padded coordinates, or actions at the same time index of independently diverging rollouts.

Let $\mathbf a^{(m)}_{t,r}$ and $\mathbf a^E_{t,r}$ be the continuous action vectors for method $m$ and task $t$'s assigned expert $\pi_{i(t)}$ on request $r$.
Use the same fixed coordinate scaling $S_a$ for all arms, derived from the expert bank's training-data action statistics, never estimated separately from each evaluated model.
The physical-to-normalized coordinate mapping, scale floor, action mask, and gripper indices are frozen in the measurement manifest.
With $D_{t,r}$ valid continuous coordinates and $N_t$ held-out requests for task $t\in\mathcal T$, report the task-macro average
\begin{equation}
 \operatorname{MSE}(m)=\frac{1}{|\mathcal T|}\sum_{t\in\mathcal T}\frac{1}{N_t}\sum_{r=1}^{N_t}
 \frac{\|S_a(\mathbf a^{(m)}_{t,r}-\mathbf a^E_{t,r})\|_2^2}{D_{t,r}}.
 \label{eq:action_mse}
\end{equation}
Discrete gripper outputs are excluded from this continuous-action metric.
Cosine similarity uses the same scaled continuous action vectors, with no model-specific recentering.
Pairs with either norm below a fixed threshold are excluded from cosine only; report their count and coverage for every arm rather than assigning artificial agreement.
All requests remain in MSE, including near-zero actions.
For the completed \pizerofive{} measurements, normalization files are identical across teachers and candidates.
We use normalized outputs with $S_a=I$, the first ten actions, six continuous coordinates, and a cosine norm threshold of $10^{-12}$.
The three final-call checkpoints yield action MSE $0.0516\pm0.0014$ and cosine similarity $0.8571\pm0.0022$; the three expert-prefix checkpoints yield $0.0655\pm0.0008$ and $0.8302\pm0.0018$, respectively.
All 400 requests have valid cosine norms for each of these six checkpoints, so no requests are excluded.
The reported standard deviations are across the three checkpoint-level task-macro scores, not across individual requests.
Each future checkpoint should pass the same normalization and native-generation checks before joining that comparison.

Reduced action MSE does not guarantee improved closed-loop success.

\subsection{Continued-Training Protocol}
\label{app:downstream}
This section details the protocol for Table~\ref{tab:continued}.
The common base, Model Soups, RegMean++, FeatCal, and TCR receive the same joint training data, task-sampling ratios, data order, optimizer, learning rate, schedule, and batch size.
Optimizer states are reset and checkpoints are evaluated on a shared grid $\mathcal B$ that includes zero and $B$.
We rerun all initializations in a common training environment and evaluate them on a shared fixed checkpoint grid. We report fixed-update scores and, separately, the best observed checkpoint on this prespecified grid as a convergence diagnostic.
The common base is the shared initialization after 200 LIBERO adaptation steps, not an untouched official pretrained checkpoint.
All initializations use one checkpoint fixed before continued training.
Each initialization is reevaluated at step zero under the continued-training protocol rather than copying its historical main-table score.
The original comparison uses evaluation seed 274001, $B=5{,}000$ optimizer updates, and the grid of zero through 5,000 in 500-update increments, with 400 episodes per checkpoint.
Continued training was extended to 10,000 updates, with the same 400-episode evaluation size at each additional checkpoint; Table~\ref{tab:continued} includes the 10,000-update score and best observed intermediate checkpoint as convergence diagnostics.
It uses rank-64 LoRA with scale parameter 64 and directly trains the action/time interfaces, with an initial learning rate of $3\times10^{-4}$, 500 warmup updates, and final learning rate $2.5\times10^{-6}$.
The training configuration specifies a global batch size of 128.

For initialization $m$, let $S_m(u)$ be success in percent after $u$ training updates. The learning-curve summary is
\begin{equation}
\operatorname{nAUC}_m=\frac{1}{B}\int_0^B S_m(u)\,du.
\label{eq:continued_metrics}
\end{equation}
The integral uses the trapezoidal rule on the common evaluation grid, with success expressed in percent; nAUC is therefore also in percent.
We compute the reported nAUC over the original 0--5,000-update window directly from the recorded checkpoint scores, without selecting the best intermediate checkpoint.
Table~\ref{tab:continued_suites} reports initial and 5,000-update success per suite to expose recovery and forgetting over the original window.
This measures practical utility as an initialization, not optimization difficulty after controlling for initial quality or end-to-end training cost.
\begin{table}[t]
\centering
\caption{Per-suite change after source multi-task training with \pizerofive{}. For initialization $m$ and suite $g$, each cell reports $S_{m,g}(0)\rightarrow S_{m,g}(B)$ and the signed change in percentage points, where $S_{m,g}(u)$ is suite success after $u$ updates. Negative changes indicate forgetting; positive changes indicate improvement. All arms share budget $B$.}
\label{tab:continued_suites}
\resulttable
\begin{tabularx}{\linewidth}{lYYYY}
\toprule
Initialization & Spatial & Object & Goal & Long \\
\midrule
Common base & $0\to92\;(+92)$ & $1\to100\;(+99)$ & $0\to93\;(+93)$ & $0\to91\;(+91)$ \\
Model Soups & $70\to91\;(+21)$ & $60\to100\;(+40)$ & $49\to94\;(+45)$ & $11\to96\;(+85)$ \\
RegMean++ & $76\to94\;(+18)$ & $85\to100\;(+15)$ & $52\to94\;(+42)$ & $22\to91\;(+69)$ \\
FeatCal & $80\to94\;(+14)$ & $88\to99\;(+11)$ & $55\to96\;(+41)$ & $39\to92\;(+53)$ \\
\rowcolor{TCRTableBlue}
\methodshort{} & $88\to97\;(+9)$ & $96\to99\;(+3)$ & $72\to97\;(+25)$ & $59\to94\;(+35)$ \\
\bottomrule
\end{tabularx}
\end{table}

\begin{figure}[t]
\centering
\includegraphics[width=\linewidth]{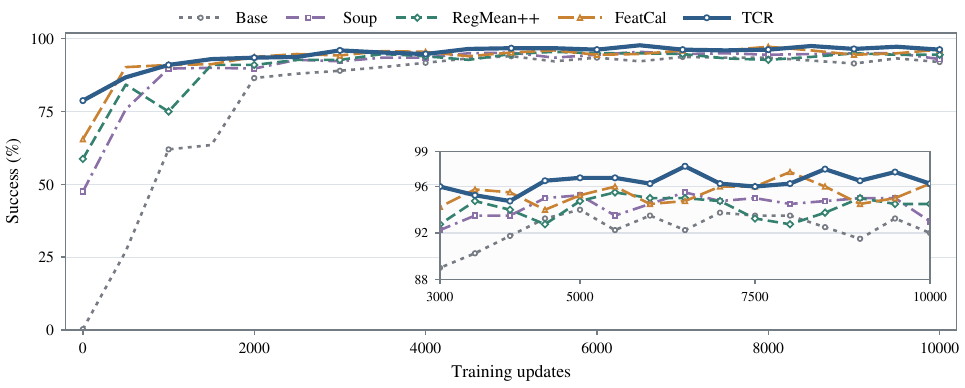}
\caption{Further training from different initializations on the same LIBERO source tasks, through 10,000 updates. Each point averages success across four suites over 400 episodes; the inset enlarges the 3,000--10,000-update phase.}
\label{fig:continued_training}
\end{figure}

\subsection{Cross-Benchmark Protocol}
The reported benchmark comparisons use \pizerofive{} on LIBERO-PRO and RoboTwin 2.0.
The PRO policies use the same four 10k-step LIBERO experts or their frozen mergers; archived 15k expert results are not substituted.
No model is trained, recalibrated, or selected on PRO.
The retained PRO comparison includes Model Soups, TIES, RegMean++, and FeatCal. All four mergers, TCR, and the preassigned expert reference have complete matched 10k first-repeat evaluations.
These benchmarks differ in expert count, embodiment, and task difficulty; their scores do not isolate any one of those factors.

Our RoboTwin 2.0 evaluation uses a 30-task subset of its official 50-task list on Aloha-AgileX, partitioned into three ten-task expert groups.
Demonstration counts, evaluation conditions, trial counts, and simulator and adapter revisions are fixed before the benchmark runs.
Each main-table column refers to one specified evaluation setting.
The expert reference executes the preassigned specialist on each task, rather than choosing an expert by test success.
Image decoding, RGB ordering, and both arms' state/action dimensions are checked against the native evaluator.
Training demonstrations, calibration traces, development states, and final resets have separate roles.
Public generalist checkpoints do not substitute for independently trained task-group experts.

\subsection{LIBERO-PRO Aggregation}
Our PRO evaluation covers position, object, language, and changed-goal perturbations. For each condition, success is macro-averaged over the 40 evaluated tasks. The PRO Average weights all four conditions equally, including Task and excluding clean LIBERO, using the same task manifest and denominators for all methods. We report the fixed first-repeat score without a cross-repeat standard deviation, as specified in Appendix~\ref{app:benchmark_task_reporting}. Per-condition results remain visible because changed-goal performance differs from same-task robustness.

\subsection{Benchmark-Specific Expert Groups and Evaluation Stages}
\label{app:benchmark_groups}
The benchmark study tests frozen-policy robustness on LIBERO-PRO and skill retention after merging a separately trained RoboTwin expert bank.
RoboTwin adds bimanual manipulation.
Neither extension requires an unseen-task or composition claim.
Expert partitions, source tasks, and evaluation conditions are fixed before training, without using comparative merger success.
Negative results remain part of the evaluation.
Continued training is studied on LIBERO.
The evaluated task lists and expert assignments are recorded separately for each benchmark.

\paragraph{LIBERO and LIBERO-PRO.}
The four LIBERO suites define four experts of ten tasks each. PRO reuses their frozen merged checkpoints and normalization/configuration without additional training, merging, trace collection for calibration, or hyperparameter selection~\citep{zhou2026liberoprorobustfairevaluation}. We record checkpoint hashes for the retained mergers. Our \pizerofive{} study reports position, object, language, and changed-goal perturbations individually and as an equally weighted four-condition average. This aggregate includes changed-goal performance, not only same-task robustness; clean LIBERO results are excluded and remain in Table~\ref{tab:main_baselines}. Independent runtime isolation prevents PRO's task registrations from modifying the clean LIBERO evaluator.

\paragraph{RoboTwin groups.}
RoboTwin does not supply LIBERO-style expert suites. Our study uses the selected Coordination, Receptacle, and Precision groups, with task assignments recorded in the expert-bank manifest. This subset is not a full 50-task benchmark result. All merging methods receive the same source experts and use the same task list and evaluation conditions. Equal task and demonstration counts do not imply equal frame counts, difficulty, or training compute; these quantities are recorded for each expert group.

\subsection{Real-Robot Protocol Details}
\label{app:real_robot}
The hardware study evaluates \pizerofive{} on a PiperX robot, using specialists fine-tuned on their respective tasks from the same base checkpoint.
All newly introduced interfaces share their initialization before fine-tuning; camera processing, action representation, control rate, and normalization conventions remain compatible across specialists.
Each merged checkpoint is evaluated under task instructions, without an expert router or post-merge training.
This experiment measures source-task retention without post-merge training.

The comparison includes each task's corresponding specialist as an empirical reference, Model Soups, TIES, RegMean++, FeatCal, and \methodshort{} (Table~\ref{tab:real_robot}).
Every merger receives the same frozen expert bank.
TIES supplies a sign-resolved, sparsified parameter-space baseline on the same materialized expert deltas~\citep{yadav2023tiesmerging}.
Its retained density and task-vector scale are fixed using development data, with the selection budget disclosed and no tuning on final hardware trials.
This is a TIES merger under the common-expert protocol, not a reproduction of the native MergeVLA system with its distinct expert construction and deployment setting~\citep{Fu_2026_CVPR}.
Calibration inputs for the hardware study are collected from physical executions of the frozen specialists.
These real-robot expert rollouts, rather than demonstration observations or simulator trajectories, are the calibration source for RegMean++, FeatCal, and TCR.
Regression methods share the recorded observation source and native input-generation convention; method-specific pass counts and realized row budgets are recorded separately.
RegMean++ uses the explicitly documented stabilized VLA adaptation, with one sequential pass and equal expert masses; it is not the two-pass uniform-weight TCR ablation or a claim of an unmodified original solver.
FeatCal starts from the uniform expert mean and computes its own forward-order feature statistics and calibration updates; it is not implemented by relabeling TCR or reusing TCR-specific weighting.
Solver, regularization, and replay differences are documented, so the hardware comparison is not interpreted as a pure weighting effect.
Calibration collection and development resets are separate from final evaluation resets.
Task~1 is plate placement and Task~2 is stacking, as identified in Table~\ref{tab:real_robot}.

\paragraph{Offline action agreement.}
Table~\ref{tab:real_robot} additionally reports overall action RMSE against the corresponding frozen specialist on recorded observations for stacking and plate placement.
The merged policy and reference expert receive identical observations and initial generation noise.
For each task, let $N_{\mathrm{req}}$ denote the number of recorded requests, $H=16$ the predicted action horizon, and $D_a=7$ the action dimension (six joint coordinates and one gripper coordinate).
The scalars $\hat a_{r,h,p}$ and $a^E_{r,h,p}$ denote merged and expert predictions at request $r$, horizon step $h$, and coordinate $p$.
We pool squared errors over requests, future steps, and action coordinates before taking the square root:
\begin{equation}
\operatorname{RMSE}_{\mathrm{expert}}
=\sqrt{\frac{1}{N_{\mathrm{req}}HD_a}\sum_{r=1}^{N_{\mathrm{req}}}\sum_{h=1}^{H}\sum_{p=1}^{D_a}
\left(\hat a_{r,h,p}-a^E_{r,h,p}\right)^2}.
\label{eq:robot_rmse}
\end{equation}
This is not the mean of per-request RMSE values, and it covers the predicted action chunk rather than only the executed portion.
The reference is the expert's prediction, not an action label from the training dataset; errors against recorded training labels measure a different quantity and are not included in this table.
Lower action RMSE indicates closer numerical agreement on the recorded inputs, not a guarantee of successful closed-loop execution.
Unlike the simulation diagnostic, this hardware metric includes the gripper and all 16 predicted steps; the simulation metric uses normalized MSE on six continuous coordinates over ten executed steps. Their numerical magnitudes are not directly comparable.

\paragraph{Trial budget.}
The reported comparison contains 20 trials per method per task, for 40 trials per method.
Each task attempt has a maximum execution time of 60\,s for every method, including the expert reference; failure to complete the task within this limit counts as a failed trial.
The five merged policies and the expert reference account for 240 reported trials in total.
The expert-reference row evaluates only the corresponding specialist on each task, for 20 trials per specialist and 40 reference trials in total; specialists are not evaluated on the other task.

\paragraph{Reporting.}
We report per-task success percentages from 20 trials and their equally weighted average.
These trials are not independent training or merging repeats.
The main table reports full-task autonomous success; any partial-progress scores are reported separately under a predefined rubric.
Timeouts, human interventions, and safety stops are logged; intervention-assisted completion is not counted as autonomous success.
Policy-induced timeouts, interventions, and safety stops count as failures, not technical-invalid exclusions.

\FloatBarrier
\section{Calibration Analysis}
\label{app:calibration_analysis}

These studies address calibration passes and cache reuse, row budgets, and other choices in the final two-pass recipe.

\subsection{Calibration Passes and Fresh Expert Data}
\label{app:cache_controls}
Does a second pass help because it repeats regression, or because it uses another set of expert execution inputs?
We separate these questions from the three component replacements in the main text.
Let A, B, and C denote expert execution caches from different episodes, collected with the same task allocation and sampling procedure.
An arrow denotes one calibration pass, not an environment rollout.
The two-pass reference is A$\to$B, using the main recipe's first-pass budget $R_1=4{,}441{,}200$ and second-pass budget $R_2=1{,}331{,}600$.

\paragraph{Reference model.}
The main TCR model also uses two-pass A$\to$B calibration; the letters denote the roles of the caches, not identical data across studies.
For these supplementary controls, we collect fresh A, B, and C caches from the same frozen expert bank and rebuild a shared first-pass A checkpoint and its A$\to$B reference.
This reference achieves 75.50\% (302/400) and is reused in Tables~\ref{tab:cache_sequence_controls}--\ref{tab:full_recipe_controls}.
The main result, $77.83\pm2.52\%$, instead summarizes three main-experiment evaluation runs.
The matching and repetition protocol is detailed in Appendix~\ref{app:ablation_protocol}.

Table~\ref{tab:cache_sequence_controls} groups the comparisons by question.
A, A$\to$A, A$\to$B, and A$\to$C share the same first-pass checkpoint.
A$\to$B$\to$B and A$\to$B$\to$C share the same two-pass checkpoint.
All second and third passes use uniform expert masses, the same row quotas, and the numerical ridge map fixed by their first-pass reference.
They recompute activations through their own merged prefixes; reusing a raw cache does not mean reusing unchanged activations.
These comparisons use the shared evaluation selection and repetition protocol in Appendix~\ref{app:ablation_protocol}.
\begin{table}[t]
\centering
\caption{Calibration passes and cache reuse. A, B, and C are distinct expert execution caches; each arrow is one pass. All rows use the same 400 evaluation episodes. A$\to$B is the shared reference for the supplementary recipe and budget studies.}
\label{tab:cache_sequence_controls}
\resulttable
\begin{tabularx}{\linewidth}{lXr}
\toprule
Calibration & Purpose & Success (\%) $\uparrow$ \\
\midrule
A & One pass & 64.75 \\
A$\to$A & Repeat the first cache & 72.75 \\
\rowcolor{TCRTableBlue}
A$\to$B & Two-pass reference with fresh data & 75.50 \\
A$\to$C & Check a second fresh cache & 78.50 \\
\midrule
A$\to$B$\to$B & Third pass, reuse B & 72.00 \\
A$\to$B$\to$C & Third pass, fresh C & \textbf{79.25} \\
\bottomrule
\end{tabularx}
\end{table}

The comparisons distinguish three effects. First, A$\to$A improves over A from 259/400 (64.75\%) to 291/400 (72.75\%): fresh episodes are not required for all of the second-pass gain.
This refinement also changes the prior, recomputed features, and weighting relative to the first pass, so it is not simply solving an unchanged regression twice.
Second, replacing the second-pass cache with B or C reaches 302/400 (75.50\%) and 314/400 (78.50\%), 2.75 and 5.75 points above A$\to$A from the same parent checkpoint.
Fresh execution inputs thus add a benefit beyond refinement on reused observations in these comparisons.
Third, more passes are not monotonically better: A$\to$B$\to$B drops to 288/400 (72.00\%), whereas A$\to$B$\to$C reaches 317/400 (79.25\%).
This pattern is consistent with repeated fitting becoming too specific to a finite cache, but does not establish overfitting as the cause; cache difficulty and representation changes remain possible explanations.
These fixed single-seed comparisons support separating cache coverage from pass count rather than treating an extra pass as an automatic improvement.

\begin{figure}[t]
\centering
\includegraphics[width=\linewidth]{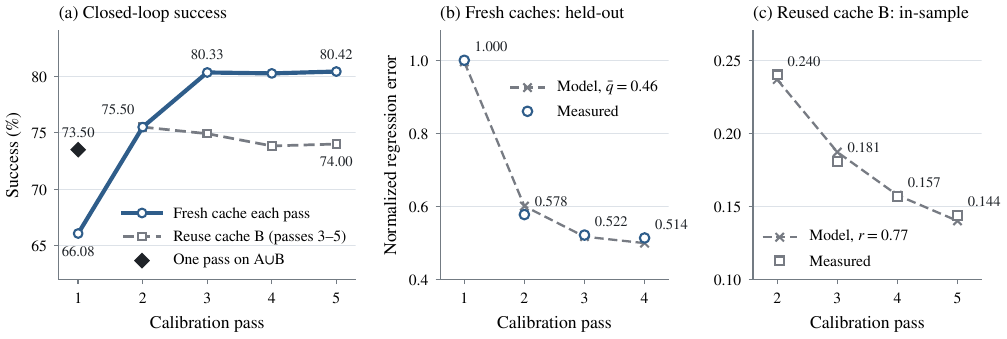}
\caption{Calibration-pass ablation on a separate calibration chain (1,200 episodes per point; 40 tasks $\times$ 30 initial states). (a) Success when every pass uses a fresh expert cache, when passes 3--5 reuse cache B, and for a single pass on the pooled cache A$\cup$B. (b) Regression error of the pass-$k$ model on the next fresh cache, normalized to pass 1, with the contraction model derived in this subsection ($\bar q$ fixed from consecutive-update cosines). (c) In-sample regression error on cache B when it is reused, with the deterministic contraction model ($r$ fixed from weight-update norms). Both error series share the same normalization. The chain is independent of Table~\ref{tab:cache_sequence_controls}; only trends are compared.}
\label{fig:calibration_passes}
\end{figure}

\paragraph{Number of calibration passes.}
To examine the pass count beyond the two-pass recipe, we run a separate calibration chain with fresh caches A--E collected from the same frozen experts and evaluate every checkpoint on 1,200 episodes (40 tasks $\times$ 30 initial states).
Figure~\ref{fig:calibration_passes} summarizes the chain.
With a fresh cache in every pass, success rises from 793/1,200 (66.08\%) after one pass to 906/1,200 (75.50\%) after two and 964/1,200 (80.33\%) after three, and then remains at 80.25\% and 80.42\% after four and five passes.
Reusing cache B in passes 3--5 instead gives 74.92\%, 73.83\%, and 74.00\%, although the in-sample regression error on B keeps decreasing (from 0.240 to 0.144).
A single pass on the pooled cache A$\cup$B with the same total rows reaches 73.50\%, compared with 75.50\% for sequential A$\to$B.
The regression error of the pass-$k$ model on the next fresh cache, normalized to pass 1, decreases from 1.000 to 0.578, 0.522, and 0.514.
Because this chain uses different caches and a larger evaluation set than Table~\ref{tab:cache_sequence_controls}, we compare only trends across the two studies.
Both show that later passes help mainly when they add new input coverage; in this chain, the gain saturates after the third pass.
The main recipe keeps two passes as a trade-off between success and construction cost.

\paragraph{Sequential prior updates.}
\label{app:sequential_prior}
In one dimension, a pass moves a weight from its prior toward the experts' target and leaves a fraction $\lambda/(a+\lambda)$ of the prior's error, where $a$ measures how strongly the cache constrains that direction; the matrix form below applies this along each eigendirection.
For one module, pass $k$ of Eq.~\eqref{eq:tcr_solve} is a prior-centered (proximal) update,
\begin{equation}
 W_k=(B_k+\lambda W_{k-1})(\Gamma_k+\lambda I)^{-1},
 \qquad B_k=\sum_i W_i\Gamma_{i,k},
 \label{eq:proximal_pass}
\end{equation}
where $\Gamma_{i,k}$ and $\Gamma_k=\sum_i\Gamma_{i,k}$ are computed from the cache and merged-prefix activations of pass $k$.
For this ridge update, before the post-solve safeguard of Eq.~\eqref{eq:tcr_safeguard}, and any fixed reference $W_\ast$, the error $E_k=W_k-W_\ast$ satisfies exactly
\begin{equation}
 E_k=E_{k-1}Q_k+R_k,
 \qquad Q_k=\lambda(\Gamma_k+\lambda I)^{-1},
 \qquad R_k=(B_k-W_\ast\Gamma_k)(\Gamma_k+\lambda I)^{-1}.
 \label{eq:pass_error}
\end{equation}
$Q_k$ contracts the error inherited from the prior, and $R_k$ is the residual of fitting the experts on the current cache.
Consider a simplified model with fixed features, a common target satisfying $B_k=W_\ast\Gamma_k$ (so $R_k=0$), simultaneously diagonalizable $\Gamma_A$ and $\Gamma_B$, and the same $\lambda$ in both settings.
A single pass on the pooled caches uses $\Gamma_A+\Gamma_B$ and multiplies the initial error by $\lambda(\Gamma_A+\Gamma_B+\lambda I)^{-1}$, whereas two sequential passes multiply it by $Q_AQ_B$.
Along a shared eigendirection with eigenvalues $a,b>0$, the contraction factors are
\begin{equation}
 q_{\mathrm{joint}}=\frac{\lambda}{a+b+\lambda},
 \qquad
 q_{\mathrm{seq}}=\frac{\lambda^2}{(a+\lambda)(b+\lambda)}.
\end{equation}
Since $(a+\lambda)(b+\lambda)=\lambda(a+b+\lambda)+ab>\lambda(a+b+\lambda)$, we have $q_{\mathrm{seq}}<q_{\mathrm{joint}}$: sequentially updating the prior removes more of the initial error than one pooled pass on the same examples.
The model omits merged-prefix recomputation, the change of weighting between passes, and the residuals $R_k$ caused by conflicting experts, so it explains the direction of the pooled-versus-sequential comparison rather than its magnitude.

\paragraph{Multiple passes.}
Unrolling Eq.~\eqref{eq:pass_error} over $K$ passes gives
\begin{equation}
 E_K=E_0\prod_{k=1}^{K}Q_k+\sum_{j=1}^{K}R_j\prod_{k=j+1}^{K}Q_k .
 \label{eq:pass_unroll}
\end{equation}
In the simplified model, pooling $K$ caches into one pass multiplies the initial error by $\lambda/(\lambda+\sum_k a_k)$ along a shared eigendirection, whereas $K$ sequential passes multiply it by $\prod_k\lambda/(\lambda+a_k)$.
Since $\prod_k(1+a_k/\lambda)>1+\sum_k a_k/\lambda$ for $K\ge2$, sequential passes contract the initial error more for any number of caches.
The residual term separates the two chains in Figure~\ref{fig:calibration_passes}.
With a fresh cache in every pass, $R_j$ changes from cache to cache.
Treating it as a fixed bias plus zero-mean cache noise and summarizing the contraction by one effective rate $\bar q$, the error follows a first-order autoregression: the inherited error decays as $\bar q^{\,k}$, the cache noise keeps a nonzero floor, and the expected held-out error approximately takes the form $b+c\,\bar q^{\,2(k-1)}$.
The same model predicts a stationary cosine of $-(1-\bar q)/2$ between consecutive updates; the measured cosines between consecutive action updates, $-0.28$ and $-0.26$, give $\bar q\approx0.46$.
With $\bar q$ fixed at this value and only $b$ and $c$ fitted, the model reproduces the held-out errors to within 0.023 (Figure~\ref{fig:calibration_passes}b), comparable to the 2\% difference observed when one model is evaluated on different fresh caches.
When cache B is reused, only the merged-prefix features change between passes, and consecutive weight updates stay aligned (cosine 0.74--0.89), so we treat $Q_k$ and $R_k$ as fixed.
The iterates then converge to the solution of $W\Gamma_B=B_B$, the unregularized least-squares fit of cache B: each further pass weakens the pull toward the prior in the directions that B covers.
The in-sample error decays at the rate of the weight updates, which shrink by a factor of about $r=0.77$ per pass; with $r$ fixed, the model reproduces the in-sample errors to within 0.007 (Figure~\ref{fig:calibration_passes}c), while success does not improve.
Both rates summarize a multi-directional contraction by a single number, and each curve has two fitted parameters, so the agreement shows that the measurements are consistent with this account rather than testing it.

\subsection{Calibration Budget}
\label{app:budget_controls}
How many regression rows are needed, and is the two-pass recipe preferable to using its row budget in one pass?
We use the same A$\to$B reference as the cache study, rather than constructing another reference for each question.
Table~\ref{tab:calibration_budget_controls} varies the second-pass row budget while holding its parent checkpoint, cache B, request coverage, expert masses, and numerical ridge fixed.
The smaller and larger settings use nested row subsets with target budgets of $0.5R_2$ and $1.5R_2$; integer rounding and small-module capacity are recorded per module.
Additional requests and duplicated rows are not used to meet a target.

The completed union control uses A and B in a single pass from Soup, with total row budget $R_1+R_2$.
It retains the single-pass relative-error weighting rule.
This compares the full methods, not just the number of passes: the two-pass version also changes its prior, the features used by later blocks, and the weighting rule.
Equal row totals do not imply equal computation; we have not measured the time needed to build these controls.
\begin{table}[t]
\centering
\caption{Row-budget comparisons. $R_1=4{,}441{,}200$ and $R_2=1{,}331{,}600$ are the reference pass budgets. Fractions denote target allocations; actual per-module counts are recorded. A$\to$B reuses the reference from Table~\ref{tab:cache_sequence_controls}, not a separate evaluation. All rows use the same 400 evaluation episodes. A dash indicates that the second pass does not apply.}
\label{tab:calibration_budget_controls}
\resulttable
\begin{tabularx}{\linewidth}{Xrrr}
\toprule
Configuration & First-pass rows & Second-pass rows & Success (\%) $\uparrow$ \\
\midrule
A$\to$B, fewer rows & $R_1$ & $0.5R_2$ & 69.00 \\
\rowcolor{TCRTableBlue}
A$\to$B, reference & $R_1$ & $R_2$ & 75.50 \\
A$\to$B, more rows & $R_1$ & $1.5R_2$ & \textbf{77.75} \\
\midrule
One pass on A$\cup$B & $R_1+R_2$ & --- & 74.25 \\
One pass on A$\cup$B$\cup$C & $R_1+2R_2$ & --- & 75.00 \\
\bottomrule
\end{tabularx}
\end{table}

The smaller and larger second-pass budgets achieve 276/400 (69.00\%) and 311/400 (77.75\%), respectively, compared with 302/400 (75.50\%) for the reference budget.
Increasing the budget by $0.5R_2$ gains 6.50 points below the reference budget and 2.25 above it, suggesting diminishing returns over the tested range rather than a linear benefit from more rows.
The one-pass A$\cup$B recipe reaches 297/400 (74.25\%), 1.25 percentage points below the two-pass reference with the same nominal row budget in this fixed single-seed comparison.
A further control pools A, B, and C for a single pass from Soup with $R_1+2R_2=7{,}104{,}400$ rows, matching the cumulative budget of A$\to$B$\to$C. It retains the union control's relative-error weighting and fixed numerical ridge and achieves 300/400 (75.00\%) on the same evaluation episodes, compared with 317/400 (79.25\%) for the three-pass recipe. This 4.25-point difference favors sequential calibration over pooling at matched row budgets; it does not isolate pass count from changes in the prior, prefix features, or weighting rule.

\paragraph{Generation-stage coverage at matched row budgets.}
The component ablation removes intermediate generator calls while also
reducing the number of regression rows. To separate generation-stage coverage
from row count, we compare multi-call and final-call-only calibration at two
matched budgets: 2.66M and 5.77M effective regression rows across two passes.
Within each budget, the two variants use the same expert execution requests and
total row count, while changing how calibration inputs are sampled from the
action-generation process.

This separate study uses three independently calibrated builds per
configuration, with different calibration seeds. Each build is evaluated on
1,200 episodes from an independent LIBERO reset bank (40 tasks with 30 resets
per task). Within each budget, the two variants are paired by calibration seed
and share the expert execution requests. At the full budget, final-call-only
calibration uses three independent initial-noise samples per fixed expert
request to match the row count, without collecting additional execution states
or retaining intermediate calls in either pass.

\begin{table}[t]
\centering
\caption{Generation-stage calibration at matched regression-row budgets
on LIBERO. Budgets count effective rows across both passes. Success is reported
as mean $\pm$ sample standard deviation over three independently calibrated
builds, each evaluated on 1,200 episodes. Gain is the difference in mean success,
in percentage points.}
\label{tab:matched_generation_budget}
\begin{tabular}{lccc}
\toprule
Total rows & Multi-call (\%) & Final-call-only (\%) & Gain (pp) \\
\midrule
2.66M & $\mathbf{77.44 \pm 1.50}$ & $74.11 \pm 1.77$ & $\mathbf{+3.33}$ \\
5.77M & $\mathbf{79.11 \pm 2.37}$ & $74.75 \pm 0.74$ & $\mathbf{+4.36}$ \\
\bottomrule
\end{tabular}%

\end{table}

Table~\ref{tab:matched_generation_budget} shows that multi-call
calibration improves mean success by 3.33 and 4.36 percentage points at the two
budgets. All six paired build comparisons favor multi-call calibration, with
gains ranging from 2.83 to 6.08 percentage points. These results support the
value of allocating calibration samples across generation stages even when
the execution requests and regression-row budget are held fixed.

\subsection{Other Choices in the Final Recipe}
\label{app:recipe_controls}
Table~\ref{tab:full_recipe_controls} tests four choices around the same two-pass reference: coverage of execution time, first-pass expert weighting, and two restrictions of the calibrated parameter scope.
Each row changes only its named rule in the construction procedure.
The early-request variant keeps the request count but uses the earliest eligible requests in each episode.
The uniform variant changes only the first pass, since the second pass is already uniform.
Both scope variants leave uncalibrated adapted parameters at the expert mean and replay through their own partially calibrated policy.
They retain Full's row quotas and numerical ridge for the modules that remain in scope; total rows and cost therefore decrease with scope.
All comparisons use the same evaluation episodes as their matching reference, following Appendix~\ref{app:ablation_protocol}.
The one-pass control is A in Table~\ref{tab:cache_sequence_controls}, not a duplicate experiment here.
\begin{table}[t]
\centering
\caption{Supplementary choices within the two-pass TCR recipe. All rows use the matching A$\to$B reference from Table~\ref{tab:cache_sequence_controls} and the same 400 episodes. Each variant changes its named rule; scope restrictions reduce the total regression budget.}
\label{tab:full_recipe_controls}
\resulttable
\begin{tabularx}{\linewidth}{lXr}
\toprule
Configuration & Change from reference & Success (\%) $\uparrow$ \\
\midrule
\rowcolor{TCRTableBlue}
Full & None; shared A$\to$B reference & \textbf{75.50} \\
Early requests only & Earliest requests, unchanged request count & 68.00 \\
Uniform first-pass weights & Equal expert masses also in pass one & 72.25 \\
Action modules only & Other adapted parameters remain at Soup & 52.25 \\
Conditioning + action & Other adapted parameters remain at Soup & 69.25 \\
\bottomrule
\end{tabularx}
\end{table}

Restricting observations to early requests reduces success from the matching Full's 75.50\% to 272/400 (68.00\%), despite preserving the request count.
This complements the main observation-source comparison: using rollout data alone is not sufficient if sampling neglects later execution states.
Uniform first-pass weights yield 289/400 (72.25\%), a 3.25-point decrease, supporting the weighting choice within this supplementary reference rather than identifying it as the source of the main method's gain.
Restricting calibration to action modules yields 209/400 (52.25\%); adding conditioning recovers 17.00 points to 277/400 (69.25\%), with Full a further 6.25 points higher.
The tested action-only correction therefore leaves substantial skill loss that broader calibration can recover. Because broader scope also increases total regression rows and computation, these are recipe comparisons, not equal-budget causal attributions to individual parameter groups.

\FloatBarrier
\section{Additional Results}
\label{app:results}

This appendix reports detailed success results, action agreement, cross-benchmark results by task group, qualitative executions, and construction and deployment cost.

\subsection{Detailed Success Results}
\label{app:detailed_success}
Table~\ref{tab:three_level_suite_results} expands all three component comparisons by suite; the matching and construction protocols are specified in Appendix~\ref{app:ablation_protocol}.
For the observation-source comparison, the main-table Full and the demonstration replacement succeed in 934 and 876 of 1,200 episodes, respectively.
Full wins 124 paired episodes and loses 66; a descriptive task bootstrap (20,000 resamples, seed 274921) gives a 95\% interval of $[-0.75,11.00]$ percentage points.
The aggregate advantage is therefore not a statistically resolved benefit across the sampled tasks.
This interval conditions on the constructed checkpoints, rather than estimating variation across independently trained expert banks.
The generation-input and prefix rows reuse their completed matched evaluations.
\begin{table}[t]
\centering
\caption{Per-suite success (\%, mean $\pm$ sample standard deviation) for the three component ablations in Table~\ref{tab:mechanism_ablation}, over three evaluations of 100 episodes per suite. Bold marks the largest displayed mean in each column.}
\label{tab:three_level_suite_results}
\resulttable
\setlength{\tabcolsep}{3pt}
\begin{tabularx}{\linewidth}{Xrrrrr}
\toprule
Configuration & Spatial & Object & Goal & Long & Average \\
\midrule
\rowcolor{TCRTableBlue}
Full & $80.00^{\scriptstyle\pm3.00}$ & $\mathbf{94.00}^{\scriptstyle\pm2.65}$ & $\mathbf{74.00}^{\scriptstyle\pm6.24}$ & $\mathbf{63.33}^{\scriptstyle\pm3.21}$ & $\mathbf{77.83}^{\scriptstyle\pm2.52}$ \\
w/o rollout obs. & $77.67^{\scriptstyle\pm0.58}$ & $90.00^{\scriptstyle\pm0.00}$ & $68.33^{\scriptstyle\pm3.21}$ & $56.00^{\scriptstyle\pm4.58}$ & $73.00^{\scriptstyle\pm1.52}$ \\
w/o intermediate calls & $79.33^{\scriptstyle\pm2.08}$ & $91.67^{\scriptstyle\pm2.52}$ & $70.67^{\scriptstyle\pm8.02}$ & $51.00^{\scriptstyle\pm5.29}$ & $73.17^{\scriptstyle\pm2.31}$ \\
w/o merged replay & $\mathbf{81.33}^{\scriptstyle\pm1.53}$ & $85.00^{\scriptstyle\pm4.36}$ & $58.00^{\scriptstyle\pm3.61}$ & $33.67^{\scriptstyle\pm4.73}$ & $64.50^{\scriptstyle\pm1.52}$ \\
\bottomrule
\end{tabularx}
\end{table}

Table~\ref{tab:libero_pro_robustness} separates PRO perturbations; benchmark task groups are reported in Appendix~\ref{app:benchmark_task_reporting}.
These are breakdowns of the main experiments, not additional method-selection sets.
\begin{table}[t]
\centering
\caption{LIBERO-PRO evaluation of frozen \pizerofive{} LIBERO checkpoints under four perturbation types: position (Pos.), object (Obj.), language (Lang.), and changed task goals (Task). Success is reported in \%. Average weights the four perturbation types equally and excludes clean LIBERO results. No PRO-driven recalibration or selection is permitted.}
\label{tab:libero_pro_robustness}
\resulttable
\begin{tabularx}{\linewidth}{lYYYYY}
\toprule
Method & Pos. & Obj. & Lang. & Task & Average \\
\midrule
\rowcolor{TCRTableGray}
Experts$^{\mathrm{ref}}$ & 18.00 & 75.75 & 91.25 & 18.00 & 50.75 \\
\midrule
Model Soups & 17.00 & 37.50 & 43.25 & \textbf{17.50} & 28.81 \\
TIES & 20.75 & 30.50 & 41.25 & 16.00 & 27.13 \\
RegMean++ & 23.00 & 49.00 & 58.00 & 11.50 & 35.38 \\
FeatCal & \textbf{30.50} & \textbf{56.75} & 63.50 & 14.75 & 41.38 \\
\midrule
\rowcolor{TCRTableBlue}
\textbf{\methodshort{} (ours)} & \textbf{30.50} & 56.50 & \textbf{75.50} & 10.25 & \textbf{43.19} \\
\bottomrule
\end{tabularx}
\end{table}

\subsection{Action Agreement}
\label{app:functional_results}
Does better task retention accompany closer agreement with the source experts?
We measure normalized action MSE and cosine on 400 shared held-out expert requests, without rerunning the closed-loop success evaluations.
Each policy generates its own complete action sequence from the same observation and initial noise.
Table~\ref{tab:main_action_agreement} compares the main methods; Table~\ref{tab:mechanism_ablation} reports the corresponding action metrics for the component study.
The measurement definitions and pairing rules are in Appendix~\ref{app:action_fidelity}.
Lower action error is a diagnostic, not a guarantee of higher task success.
\begin{table}[t]
\centering
\caption{Action agreement for the main LIBERO methods on 400 shared held-out expert requests. Success is copied from Table~\ref{tab:main_baselines}; no additional success evaluation is used here. Action metrics use the same request bank.}
\label{tab:main_action_agreement}
\resulttable
\begin{tabularx}{\linewidth}{Xrrr}
\toprule
Method & Action MSE $\downarrow$ & Action cos $\uparrow$ & Success (\%) $\uparrow$ \\
\midrule
Model Soups & 0.1094 & 0.7547 & 44.83 \\
Task Arithmetic & 0.1196 & 0.7516 & 46.00 \\
TIES-Merging & 0.1169 & 0.7452 & 44.25 \\
WUDI-Merging & 0.1218 & 0.7427 & 46.00 \\
KnOTS-TIES & 0.0980 & 0.7661 & 47.50 \\
RegMean & 0.0881 & 0.7899 & 53.67 \\
RegMean++ & 0.0773 & 0.8080 & 58.17 \\
FeatCal & 0.0678 & 0.8285 & 68.33 \\
\midrule
\rowcolor{TCRTableBlue}
\textbf{TCR} & $\mathbf{0.0488}^{\scriptstyle\pm 0.0011}$ & $\mathbf{0.8620}^{\scriptstyle\pm 0.0017}$ & \textbf{77.83} \\
\bottomrule
\end{tabularx}
\par\smallskip
\begin{minipage}{\linewidth}
\footnotesize
The RegMean row uses the same frozen $\alpha=0.3$ checkpoint for action agreement and the success rate copied from Table~\ref{tab:main_baselines}.
\end{minipage}
\end{table}

Across RegMean, RegMean++, FeatCal, and TCR, MSE decreases from 0.0881 to 0.0488 while success increases from 53.67\% to 77.83\%; cosine follows the same ordering.
This association is descriptive across the evaluated constructions, not an independently controlled effect of action error.
It is not a strict ranking rule even within simulation: Soup has lower MSE than Task Arithmetic (0.1094 versus 0.1196), but lower success (44.83\% versus 46.00\%).
The physical plate-placement reversal in Section~\ref{sec:real_robot} provides another counterexample; hardware RMSE and simulation MSE use different action representations and should not be compared numerically across domains.
These observations motivate reporting functional fidelity alongside, rather than instead of, closed-loop success.

\paragraph{Compromise toward other experts.}
Action error does not show whether a merged action drifts toward the behavior of the other experts. We therefore project the deviation of the merged action from the owner expert, the specialist for the current task, onto the direction of the other experts' mean action:
\begin{equation}
t=\frac{(a_{\mathrm{merge}}-a_{\mathrm{owner}})^\top(\bar a_{\mathrm{other}}-a_{\mathrm{owner}})}{\lVert\bar a_{\mathrm{other}}-a_{\mathrm{owner}}\rVert_2^2},
\qquad
\bar a_{\mathrm{other}}=\frac{1}{3}\sum_{j\neq\mathrm{owner}}a_j .
\label{eq:compromise_index}
\end{equation}
Here $t=0$ indicates no shift toward the other experts, and $t=0.75$ corresponds to the equally weighted average of the four expert actions.
We evaluate $t$ on 400 states drawn from the owner experts' own executions, 100 per suite, with the same generation noise for all policies.
Table~\ref{tab:compromise_index} reports $t$ together with wrong-target selections in closed-loop Goal-suite evaluation (100 episodes for each of the five policies and three builds, 1,500 episodes in total); an episode counts as a wrong-target selection when it completes another Goal task or moves only objects not named in the instruction.
On the 308 steps where the owner expert and the other experts command opposite gripper directions, TCR produces no intermediate gripper value.
The remaining TCR failures on the Goal suite involve the instructed object and arise from execution.
\begin{table}[t]
\centering
\caption{Compromise toward the other experts on LIBERO. The index $t$ (Eq.~\ref{eq:compromise_index}) is averaged over 400 owner-expert execution states with shared generation noise; $t=0$ follows the owner expert and $t=0.75$ is the four-expert average. All three TCR builds give $t=0.34$. Wrong-target selections are counted per 100 closed-loop Goal-suite episodes, pooled over three builds. The expert row resamples the owner expert's generation noise.}
\label{tab:compromise_index}
\resulttable
\begin{tabularx}{\linewidth}{Xrr}
\toprule
Method & Compromise index $t$ $\downarrow$ & Wrong targets per 100 episodes $\downarrow$ \\
\midrule
Owner expert & 0.01 & 0.0 \\
\midrule
Model Soups & 0.60 & 15.7 \\
RegMean++ & 0.47 & 12.7 \\
FeatCal & 0.37 & 5.3 \\
\midrule
\rowcolor{TCRTableBlue}
\textbf{TCR} & \textbf{0.34} & \textbf{0.3} \\
\bottomrule
\end{tabularx}
\end{table}

\begin{figure}[t]
\centering
\includegraphics[width=\linewidth]{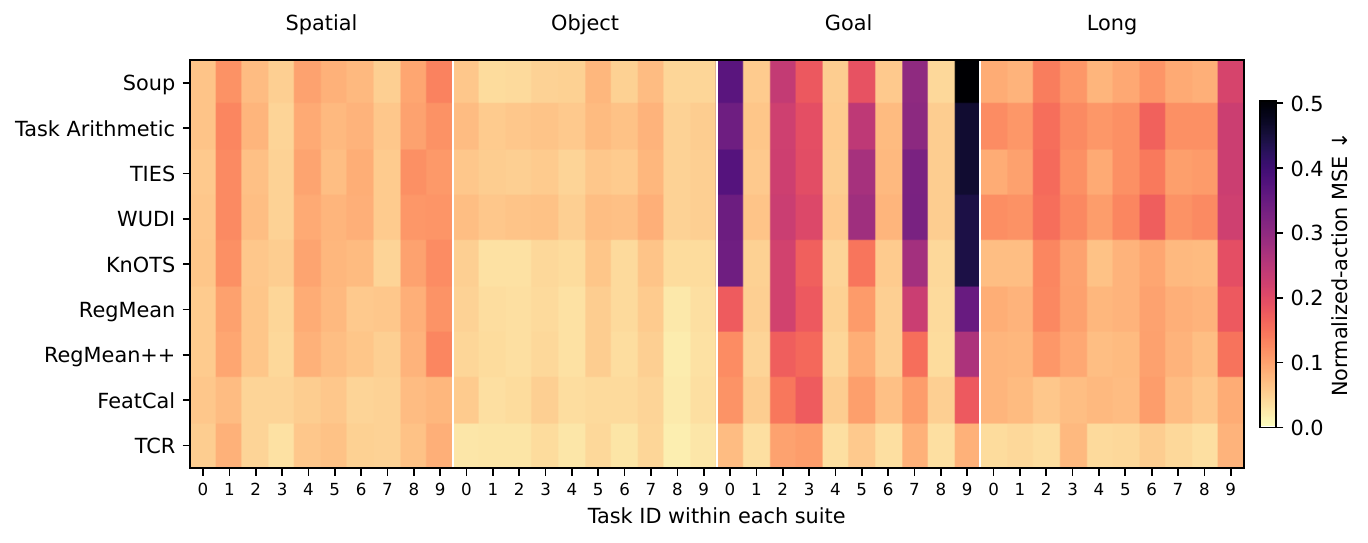}
\par\smallskip
\includegraphics[width=\linewidth]{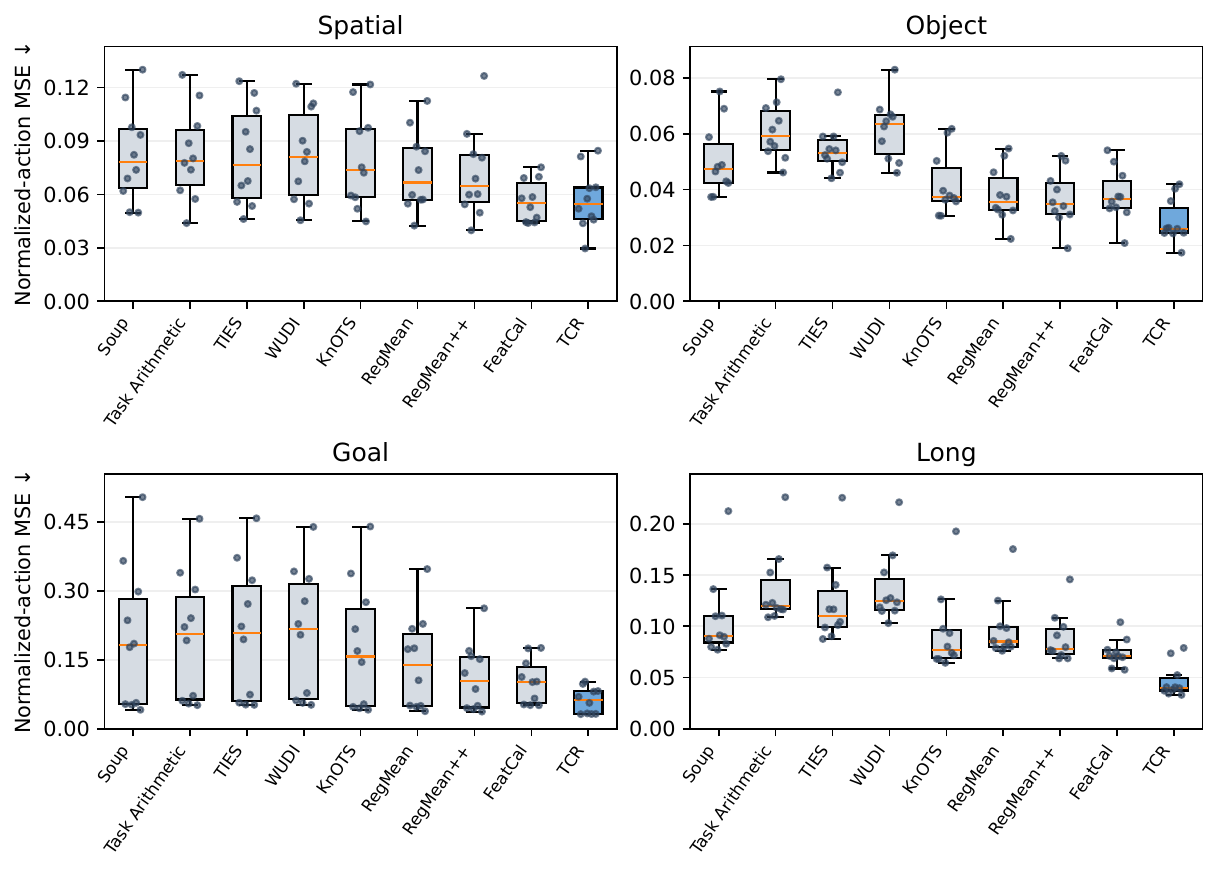}
\caption{Task-resolved normalized-action error for nine \pizerofive{} merging methods on the common held-out request bank. Top: all 40 task means with one shared color scale (lower is better). Bottom: the ten task means within each suite, with suite-specific vertical-axis ranges starting at zero; points retain every task, while boxes summarize their distribution, not uncertainty across seeds. Predictions are compared with the assigned expert on identical observations and initial noise. Expert self-error is identically zero and is omitted.}
\label{fig:task_action_fidelity}
\end{figure}

\paragraph{Task-resolved action fidelity.}
Figure~\ref{fig:task_action_fidelity} expands the held-out comparison across all nine merging methods and all 40 tasks.
Each cell or point averages the same ten requests for one task; the visualization uses the first evaluated policy for each method.
The shared color scale exposes task-specific differences that an overall average can hide.
For example, the Goal-suite MSE is 0.1969 for Model Soups, 0.1031 for FeatCal, 0.1123 for RegMean++, and 0.0618 for TCR.
The ordering is not uniform: FeatCal has slightly lower Spatial-suite MSE than TCR (0.0563 versus 0.0569).
These observations describe output agreement on expert observations, not neuron activation or causal explanations of closed-loop success.

\subsection{Cross-Benchmark Results by Task Group}
\label{app:benchmark_task_reporting}
The main table reports one task-averaged score per benchmark and method.
Table~\ref{tab:cross_benchmark_groups} gives the group breakdowns, and Table~\ref{tab:libero_pro_robustness} separates PRO perturbation types.
Task IDs, group membership, demonstration counts, expert assignments, and evaluation conditions accompany the final task list.

Let $s_{m,k,t}$ be method $m$'s success percentage on task $t$ in evaluation repeat $k$.
For disjoint task groups $\mathcal T_g$ with $\mathcal T=\bigcup_g\mathcal T_g$, we compute
\begin{equation}
 S_{m,k,g}=\frac{1}{|\mathcal T_g|}\sum_{t\in\mathcal T_g}s_{m,k,t},
 \qquad
 S_{m,k}=\frac{1}{|\mathcal T|}\sum_g|\mathcal T_g|S_{m,k,g}.
 \label{eq:group_success}
\end{equation}
The expert reference substitutes the score of task $t$'s preassigned specialist for $s_{m,k,t}$.
It is an empirical reference, not an upper bound or a single generalist policy.
All other method columns use one merged checkpoint per repeat across the complete task set.

The main cross-benchmark tables report one fixed evaluation repeat
(the first) for every method, without selecting the best-scoring repeat.
A supplementary three-run evaluation of the specialists, RegMean++, and TCR
on RoboTwin is reported below.

LIBERO-PRO uses evaluation seed 274001 and its frozen perturbation/reset selections: ten episodes on each of 40 tasks under each of four perturbation types, totaling 1,600 episodes per method.
RoboTwin uses the first repeat's task-specific seed/reset manifests: six episodes on each of 30 tasks, totaling 180 episodes per method; a single repeat does not mean using one identical environment seed for all episodes.
The expert bank, hyperparameters, and corresponding first-repeat checkpoints remain fixed, and methods use paired task resets and evaluation seeds.
Group and overall scores are computed from this complete repeat only; these results do not estimate variation across repeats or expert-training runs.

\begin{table}[t]
\centering
\caption{Group-level cross-benchmark success with \pizerofive{} (\%). LIBERO-PRO groups tasks by their source LIBERO suite; RoboTwin uses three preassigned expert groups. The overall score weights source tasks equally.}
\label{tab:cross_benchmark_groups}
\resulttable
\setlength{\tabcolsep}{4pt}
\begin{tabularx}{\linewidth}{lYYYYY}
\toprule
\multicolumn{6}{l}{\textbf{(a) LIBERO-PRO}} \\
\addlinespace[2pt]
\textbf{Method} & \textbf{Spatial} & \textbf{Object} & \textbf{Goal} & \textbf{Long} & \textbf{Average} $\boldsymbol{\uparrow}$ \\
\midrule
\rowcolor{TCRTableGray}
Experts$^{\mathrm{ref}}$ & 54.75 & 56.75 & 46.75 & 44.75 & 50.75 \\
\midrule
Model Soups & 45.75 & 35.25 & 30.25 & 4.00 & 28.81 \\
TIES & 46.25 & 34.50 & 19.50 & 8.25 & 27.13 \\
RegMean++ & 45.00 & 52.50 & 36.50 & 7.50 & 35.38 \\
FeatCal & \textbf{56.25} & \textbf{58.25} & 34.00 & 17.00 & 41.38 \\
\midrule
\rowcolor{TCRTableBlue}
\textbf{\methodshort{}} & 55.25 & 56.25 & \textbf{40.00} & \textbf{21.25} & \textbf{43.19} \\
\bottomrule
\end{tabularx}
\vspace{5pt}
\begin{tabularx}{\linewidth}{lYYYY}
\toprule
\multicolumn{5}{l}{\textbf{(b) RoboTwin 2.0}} \\
\addlinespace[2pt]
\textbf{Method} & \textbf{Coordination} & \textbf{Receptacle} & \textbf{Precision} & \textbf{Average} $\boldsymbol{\uparrow}$ \\
\midrule
\rowcolor{TCRTableGray}
Experts$^{\mathrm{ref}}$ & 60.00 & 56.67 & 56.67 & 57.78 \\
\midrule
Model Soups & 3.33 & 20.00 & \textbf{13.33} & 12.22 \\
TIES & 3.33 & 10.00 & 5.00 & 6.11 \\
RegMean++ & 3.33 & 25.00 & 11.67 & 13.33 \\
FeatCal & 0.00 & 0.00 & 0.00 & 0.00 \\
\midrule
\rowcolor{TCRTableBlue}
\textbf{\methodshort{}} & \textbf{5.00} & \textbf{33.33} & 8.33 & \textbf{15.56} \\
\bottomrule
\end{tabularx}
\end{table}

\paragraph{FeatCal on RoboTwin.}
\label{app:featcal_robotwin_diagnostic}
The evaluated FeatCal adaptation uses static observations with three noise replicas, all at $\tau_{\mathrm{flow}}=1$. Its two time-MLP input caches consequently have rank one: the 450 calibration rows repeat the same 1024-dimensional time input. The FeatCal update in the unobserved complement is governed by its anchor rather than by feature matches at later flow times. Relative to the initial Model Soup, the time-conditioning vector changes by 3.28\% at $\tau_{\mathrm{flow}}=1$ but 90.35\% at $\tau_{\mathrm{flow}}=0.6$. In the 180 paired RoboTwin episodes, FeatCal has a systematic first-action offset relative to Soup and RegMean++, and several failed trajectories show absent or delayed task-relevant gripper closure; this behavior is not universal across failures.

To probe the time path, we restored only the two time-MLP weights from Soup in memory, leaving the other 811 state tensors unchanged, and ran native action generation on nine fixed calibration observations spanning the three task groups. The mean L2 distance of the first normalized action to its expert fell from 0.1765 to 0.1052; executed arm-target and gripper-command errors also decreased. This single-factor offline intervention shows that time-path drift contributes to action error. The nine observations are not held out, and no closed-loop evaluation of this diagnostic variant is used here; neither the intervention nor the trajectory inspection establishes that this drift alone caused FeatCal's 0/180 success.

\paragraph{Absolute success on LIBERO-PRO.}
Under the evaluated distribution shifts, the specialists themselves achieve 50.75\% success.
TCR achieves 43.19\%, retaining approximately 85.1\% of specialist performance, the highest among the evaluated merging methods.

\paragraph{Repeated evaluation on RoboTwin.}
We additionally evaluate the specialists, RegMean++, and TCR in three
runs each. Each run contains 180 episodes across 30 tasks, giving 540 episodes
per method, with task and reset assignments matched across methods within
each run. Table~\ref{tab:robotwin_repeated_eval} reports the individual runs
and their mean and sample standard deviation. These are evaluation repeats,
not independent expert-training runs.

\begin{table}[t]
\centering
\caption{Repeated RoboTwin evaluation. Each run evaluates 30 tasks
with six episodes per task, for 180 episodes per run and 540 per method.
All entries are success rates in percent; the final column reports mean
$\pm$ sample standard deviation across the three runs.}
\label{tab:robotwin_repeated_eval}
\begin{tabular}{lcccc}
\toprule
Method & Run 1 & Run 2 & Run 3 & Mean $\pm$ SD \\
\midrule
Specialists & 57.78 & 56.11 & 56.67 & $56.85 \pm 0.85$ \\
RegMean++ & 13.33 & 11.11 & 11.67 & $12.04 \pm 1.16$ \\
\textbf{TCR} & \textbf{15.56} & \textbf{20.56} & \textbf{16.67}
& $\mathbf{17.59 \pm 2.63}$ \\
\bottomrule
\end{tabular}%

\end{table}

TCR outperforms RegMean++ in all three runs, with a mean gain of
5.56 percentage points computed from the unrounded success rates. The lower
bound of the task-cluster interval for this difference is zero. Although the
ordering is consistent across runs, TCR's mean success of 17.59\% remains
substantially below the specialists' 56.85\%, indicating that substantial
skill degradation remains on this benchmark.

\paragraph{Failure modes on RoboTwin.}
We further replay 1,246 stored trajectories and classify failures
using their physical states. In this diagnostic set, TCR has 92 more grasp
failures and 79 more contact/precision failures than the specialists,
compared with 18 more placement failures. Among these categories, the largest
increases occur in grasping and contact execution, identifying concrete
manipulation skills that remain vulnerable after merging.

\paragraph{Comparing benchmarks.}
The benchmarks use different task partitions and expert counts. These comparisons also change data, embodiment, and difficulty, so they do not establish scaling with expert count.

Inference latency, memory, and checkpoint size use the shared hardware protocol in Appendix~\ref{app:cost}.

\subsection{Qualitative Execution Sequences}
\label{app:qualitative_rollouts}
\begin{figure}[t]
\centering
\includegraphics[width=\linewidth]{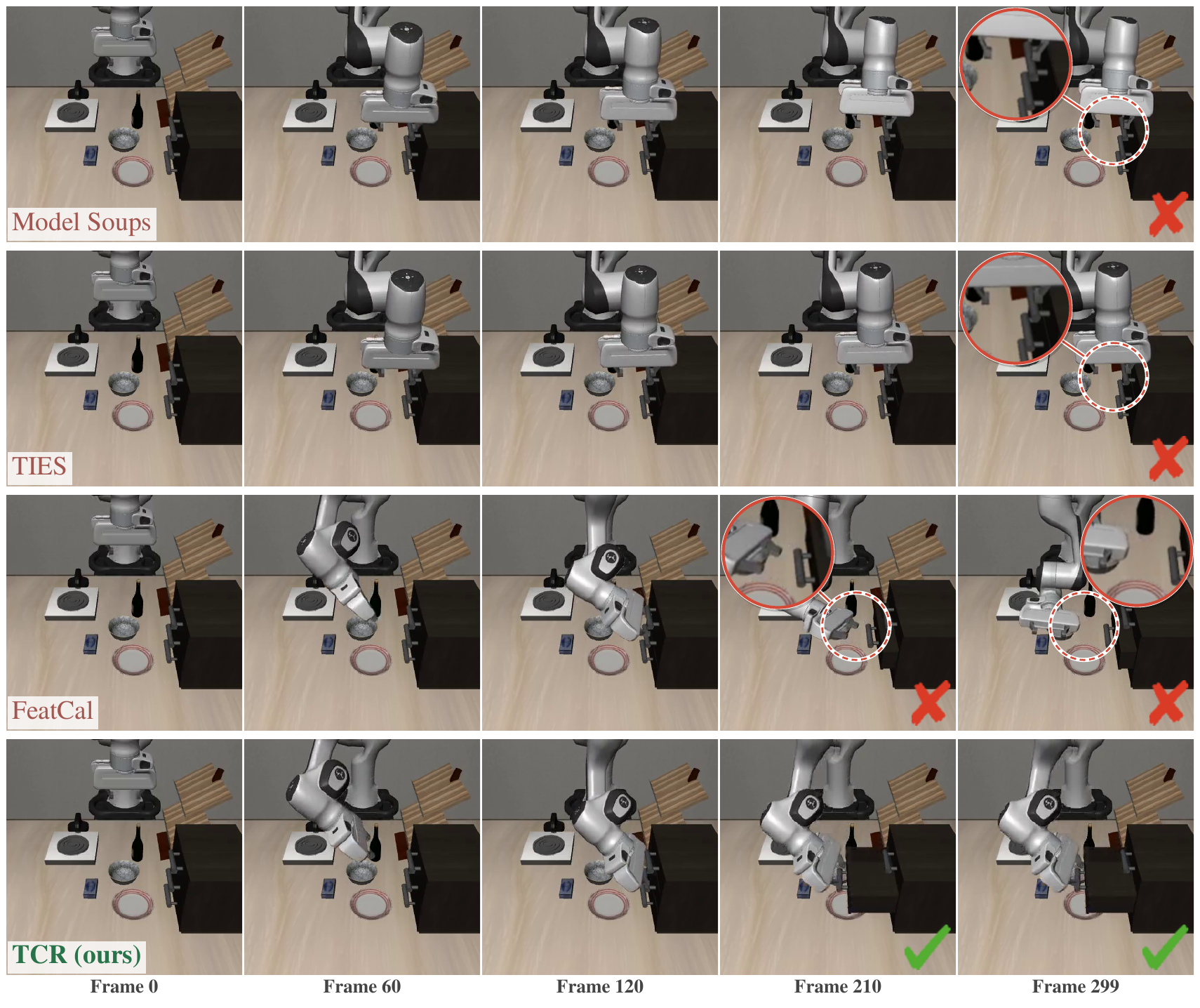}
\caption{\textbf{Skill retention during execution.} Four merged policies attempt to open the middle drawer from the same initial state. Columns show shared recorded-frame indices 0, 60, 120, 210, and 299. TCR succeeds at its final recorded frame 140; that terminal image is held in the 210 and 299 columns, both marked with green checks. Red crosses mark FeatCal's visible wrong-drawer state at 210 and 299 and the terminal failures of Model Soups and TIES at 299. Circular insets show both the gripper and drawer handles, preserving their spatial relationship in the same frames. These marks identify sampled outcomes, not exact failure-onset times. This selected example illustrates behavior, not aggregate success.}
\label{fig:qualitative_rollout_1}
\vspace{-15pt}
\end{figure}

\begin{figure}[t]
\centering
\includegraphics[width=\linewidth]{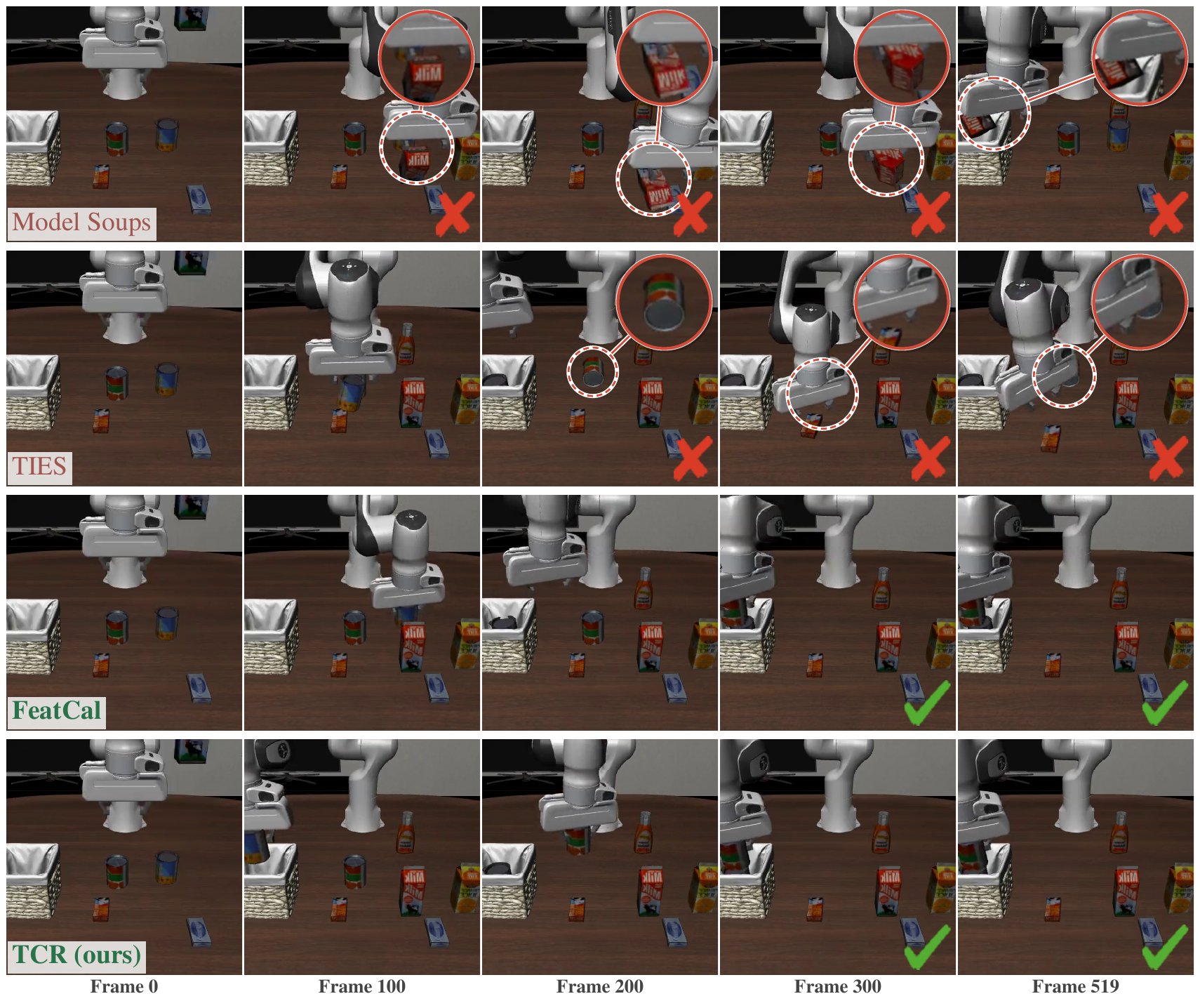}
\caption{Place soup and tomato sauce in the basket from a matched initial simulator state. Columns show recorded frame indices, not seconds. TCR and FeatCal succeed; Model Soups and TIES fail. Checks mark held terminal frames of successful rollouts; crosses mark visible errors at sampled frames, enlarged in circular insets. These selected examples illustrate behavior rather than estimate success rates.}
\label{fig:qualitative_rollout_2}
\end{figure}

Figures~\ref{fig:qualitative_rollout_1} and~\ref{fig:qualitative_rollout_2} show selected drawer-opening and basket-placement episodes, comparing TCR, Model Soups, TIES, and FeatCal.
Within each example, task identity and the initial simulator state are matched across methods.
Frames are sampled at common recorded-frame indices without aligning task milestones or rescaling each policy's trajectory duration.
The examples illustrate differences in execution; the aggregate comparison remains Table~\ref{tab:main_baselines}.

\subsection{Construction and Deployment Cost}
\label{app:cost}
Table~\ref{tab:cost_build} reports construction time and sampled peak GPU memory for the reported checkpoints.
Expert training and validation screening are excluded from merge times; data collection is included where applicable.

TCR uses 4,441,200 first-pass and 1,331,600 second-pass regression rows, totaling 5,772,800.
\begin{table}[h]
\centering
\caption{\pizerofive{} construction accounting for the main comparison. States counts request-flow inputs; Demo. counts supervised states.}
\label{tab:cost_build}
\footnotesize
\renewcommand{\arraystretch}{1.12}
\setlength{\tabcolsep}{4pt}
\begin{tabularx}{\linewidth}{Xrrrrrr}
\toprule
Method & States & Rows & Demo. & Grad. steps & Merge (h) & Peak GiB \\
\midrule
Model Soups & 0 & 0 & 0 & 0 & 0.04 & $<0.1$ \\
Task Arithmetic (val.) & 0 & 0 & 0 & 0 & 0.09 & 0 \\
TIES-Merging (val.) & 0 & 0 & 0 & 0 & 0.25 & 0 \\
WUDI-Merging & 0 & 0 & 0 & $300^{\dagger}$ & 0.30 & 2.4 \\
KnOTS-TIES & 0 & 0 & 0 & 0 & 0.03 & 0.9 \\
RegMean & 200 & $2.37$M & 0 & 0 & 0.36 & 21.3 \\
RegMean++ & 200 & $2.37$M & 0 & 0 & 0.35 & 21.3 \\
FeatCal & 600 & $4.44$M & 0 & 0 & 1.55 & 20.5 \\
\midrule
\rowcolor{TCRTableBlue}
\textbf{\methodshort{} (ours)} & 1,200 & $5.77$M & 0 & 0 & 1.03 & 32.8 \\
\bottomrule
\end{tabularx}
\par\smallskip
\footnotesize $^{\dagger}$WUDI steps are per optimized matrix. TCR timing and peak-memory scope are detailed below.
\end{table}

Task Arithmetic and TIES construction times profile only their selected checkpoints after screening. Both materialize on CPU with CUDA hidden, so peak GPU usage is zero. Model Soups was re-materialized to the reported checkpoint in 155.74 seconds, with 5 MiB sampled whole-device usage on an otherwise empty A100.

RegMean and RegMean++ each select 50 single-flow inputs per expert; FeatCal uses 50 observations with three noise replicas per expert, all at flow time $t=1$. The regression times cover four source captures and solve/export; FeatCal additionally includes input preparation and parallel source capture. Row totals are rounded from the construction records.

TCR's merge time sums compatible historical stages rather than timing a full build. Its 32.8~GiB peak is the process peak of the two-pass build, observed consistently in two independent measurements, one on an exclusive GPU and one with per-process sampling on a shared GPU. FeatCal's peak is measured per process and is reached when two source captures run concurrently. TCR's two-pass build uses more memory than FeatCal; its advantage lies in construction time. Other GPU peaks use one-second whole-device sampling.

\begin{figure}[t]
\centering
\includegraphics[width=\linewidth]{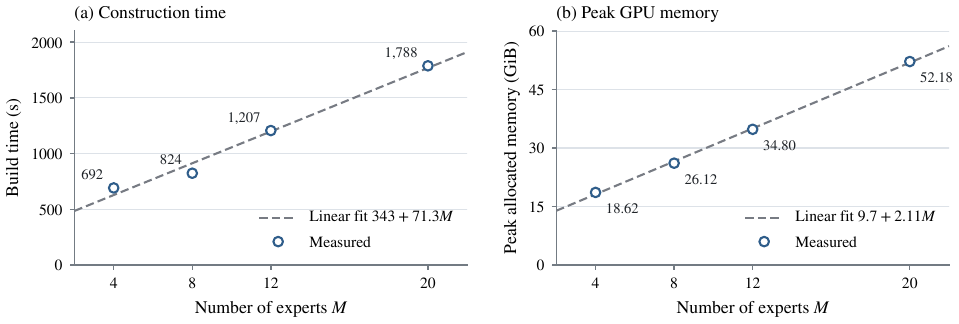}
\caption{One-pass construction cost versus the number of experts on one A100-80GB. Banks with $M>4$ replicate the four experts (cost only; success is not evaluated). (a) Build time. (b) Peak allocated GPU memory. Dashed lines are least-squares linear fits.}
\label{fig:cost_scaling}
\end{figure}

\paragraph{Scaling with the number of experts.}
To measure how construction cost grows with the number of experts, we profile a single first pass (pass A) on banks in which experts beyond the four are replicas; this study measures cost only and does not evaluate success (Figure~\ref{fig:cost_scaling}).
The build streams the merged-prefix K/V from host memory one layer at a time, since each block reads only its own layer, and solves Eq.~\eqref{eq:tcr_solve} in its $d_{\mathrm{in}}\times d_{\mathrm{in}}$ form, whose system size does not grow with the number of rows.
Rows grow exactly linearly in the number of experts (4.44M to 22.21M for $M=4$ to 20). On an exclusive A100-80GB, peak allocated memory follows $P(M)\approx9.7+2.11M$~GiB, about 2.11~GiB per additional expert, and a least-squares fit of the build time gives $T(M)\approx343+71.3M$~s, about 1.2 minutes per additional expert, from 692~s at $M=4$ to 1,788~s at $M=20$; individual increments are not uniform because weight loading and I/O vary between runs.
In a separately profiled four-expert build of 711~s, the 418 module solves take 178~s, feature capture and merged-prefix replay 184~s, and loading weights, validating replay references, and writing the 9.35~GB checkpoint about 349~s.

\paragraph{Deployment.}
\methodshort{} deploys one checkpoint covering all four skill families, whereas the expert bank stores four checkpoints of the same architecture.
The fused policy retains the expert architecture and native inference procedure, without an additional router or ensemble execution.
Thus, checkpoint storage is reduced to approximately one quarter of the full expert bank; per-request inference cost remains comparable to running one expert, with no claimed structural speedup.

\FloatBarrier

\section{Limits of Embodied Policy Mergeability}
\label{app:mergeability}

This appendix asks when independently fine-tuned embodied policies can be merged into one set of weights without routing while retaining their closed-loop skills. Section~\ref{app:conflict_bound} gives a conditional bound for a shared linear module, Section~\ref{app:limits_count} measures retention as the expert set grows, and Section~\ref{app:limits_fidelity} relates pre- and post-merge measurements to retention. All LIBERO results in this appendix use the four $\pi_{0.5}$ suite specialists, the procedural reset bank of the main evaluation, and three evaluation repeats per model and suite. Retention is the merged model's success count divided by the specialist's on the same suite and repeat.

\subsection{A Conditional Conflict Bound}
\label{app:conflict_bound}

Consider one linear module whose input features are held fixed. Expert $i\in\{1,\dots,M\}$ has weights $W_i$ and input matrix $X_i$ with $n_i$ rows, and $\Sigma_i=X_i^\top X_i/n_i$. A shared weight matrix $W$ has average output-matching error
\[
\mathcal E_M(W)=\frac1M\sum_{i=1}^M\frac1{n_i}\big\|X_i(W-W_i)^\top\big\|_F^2
=\frac1M\sum_{i=1}^M\operatorname{tr}\!\big((W-W_i)\Sigma_i(W-W_i)^\top\big).
\]

\paragraph{Proposition 1 (fixed-feature conflict bound).}
Suppose there is a positive semidefinite matrix $Q$ and constants $\gamma,\delta>0$ such that $\Sigma_i\succeq\gamma Q$ for every $i$ and $\|(W_i-W_j)Q^{1/2}\|_F^2\ge\delta^2$ for every $i\ne j$. Then, for every $W$,
\begin{equation}
\mathcal E_M(W)\;\ge\;\frac{\gamma\delta^2}{2}\Big(1-\frac1M\Big).
\label{eq:conflict_bound}
\end{equation}

\emph{Proof.} Let $Z_i=W_iQ^{1/2}$, $Y=WQ^{1/2}$, and $\bar Z=\frac1M\sum_iZ_i$. Since $\Sigma_i\succeq\gamma Q$, each term satisfies $\operatorname{tr}((W-W_i)\Sigma_i(W-W_i)^\top)\ge\gamma\|Y-Z_i\|_F^2$. The mean minimizes the sum of squared distances, and $\sum_i\|Z_i-\bar Z\|_F^2=\frac1M\sum_{i<j}\|Z_i-Z_j\|_F^2$. Hence
\[
\mathcal E_M(W)\ge\frac{\gamma}{M}\sum_i\|Y-Z_i\|_F^2\ge\frac{\gamma}{M}\sum_i\|\bar Z-Z_i\|_F^2=\frac{\gamma}{M^2}\sum_{i<j}\|Z_i-Z_j\|_F^2\ge\frac{\gamma}{M^2}\cdot\frac{M(M-1)}{2}\,\delta^2,
\]
which equals the right-hand side of Eq.~\eqref{eq:conflict_bound}. $\square$

\paragraph{Corollary 1 (number of experts within a tolerance).}
Under the assumptions of Proposition~1, if a shared module must satisfy $\mathcal E_M(W)\le\varepsilon^2$ and $\gamma\delta^2>2\varepsilon^2$, then $M\le\big\lfloor\gamma\delta^2/(\gamma\delta^2-2\varepsilon^2)\big\rfloor$. This follows by solving Eq.~\eqref{eq:conflict_bound} for $M$.

\emph{Remarks.} With expert weights $\omega_i\ge0$, $\sum_i\omega_i=1$, as in Eq.~\eqref{eq:tcr_objective}, the same argument applied to $\sum_i\omega_i\operatorname{tr}((W-W_i)\Sigma_i(W-W_i)^\top)$ gives the bound $\frac{\gamma\delta^2}{2}(1-\sum_i\omega_i^2)$, which reduces to Eq.~\eqref{eq:conflict_bound} for uniform weights. The ridge term of the TCR objective is nonnegative, so the bound also applies to the matching part of TCR's solution. The result is conditional: it concerns module-level matching with fixed features, we do not verify that the assumed common component $Q$ exists in the trained modules, and it does not bound closed-loop success. It states that when experts use commonly activated feature directions differently, no single shared linear map can match all of them on those directions.

\subsection{Number of Merged Experts}
\label{app:limits_count}

We build TCR and Model Soups on all 11 subsets of the four specialists: six pairs, four triples, and the full set. All TCR subset builds use the two-pass recipe of the main method with the same settings; the number of calibration rows per expert is fixed, so the total grows linearly with the number of experts. Each merged model is evaluated on every suite it contains, with 10 tasks $\times$ 10 episodes per suite and three repeats, i.e., 300 episodes per model and suite. For each suite, we average retention over all subsets that contain it, so that the result does not depend on one particular ordering of experts.

\begin{figure}[t]
\centering
\includegraphics[width=0.5\linewidth]{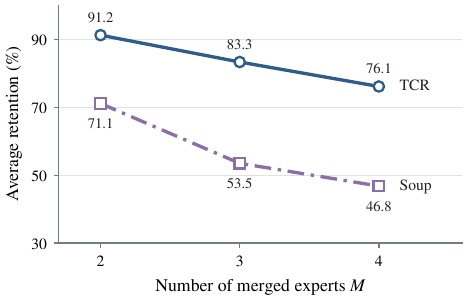}
\caption{Average retention over the four LIBERO suites as the number of merged experts grows. Each point averages, per suite, all subsets of the four specialists that contain that suite. Per-suite values are in Table~\ref{tab:limits_expert_count}.}
\label{fig:limits_expert_count}
\end{figure}

\begin{table}[t]
\centering
\caption{Retention (\%) by suite and number of merged experts $M$, averaged over all subsets containing the suite.}
\label{tab:limits_expert_count}
\resulttable
\begin{tabular}{lcccccc}
\toprule
& \multicolumn{3}{c}{TCR} & \multicolumn{3}{c}{Model Soups} \\
\cmidrule(lr){2-4}\cmidrule(lr){5-7}
Suite & $M=2$ & $M=3$ & $M=4$ & $M=2$ & $M=3$ & $M=4$ \\
\midrule
Spatial & 92.0 & 86.5 & 81.5 & 77.8 & 72.9 & 74.3 \\
Object & 99.3 & 93.3 & 89.9 & 90.8 & 76.0 & 54.7 \\
Goal & 94.6 & 80.9 & 68.8 & 69.2 & 45.9 & 47.6 \\
Long & 78.8 & 72.6 & 64.3 & 46.6 & 19.0 & 10.7 \\
\midrule
Average & \textbf{91.2} & \textbf{83.3} & \textbf{76.1} & 71.1 & 53.5 & 46.8 \\
\bottomrule
\end{tabular}
\end{table}

Retention decreases by about 7.5 percentage points per added expert, with no abrupt collapse in the tested range (Figure~\ref{fig:limits_expert_count}). The margin over Model Soups grows from 20.1 percentage points with two experts to 29.9 and 29.3 with three and four. Difficulty depends mainly on the suite: Object retains 89.9\% with four experts, whereas Goal and Long drop to 68.8\% and 64.3\%. On the fixed Spatial$\to$Goal chain evaluated on Spatial and Goal (600 episodes per point), TCR reaches 87.2\%, 80.5\%, and 71.0\% for two, three, and four experts; FeatCal reaches 76.2\%, 80.5\%, and 68.2\%, so the two methods are tied at three experts.

Three qualifications apply. First, LIBERO has four suites, so these results do not identify a maximum number of mergeable experts beyond four. Second, the four-expert TCR point comes from the same subset study as the smaller subsets and is about 4 percentage points below the main-table checkpoint. Third, subsets differ in expert composition and total calibration rows, so the decline is an empirical property of the current recipe as the expert set grows rather than an effect of expert count alone. The earlier fixed nested-subset study with single evaluations is also reported below.

\begin{figure}[t]
\centering
\includegraphics[width=0.76\linewidth]{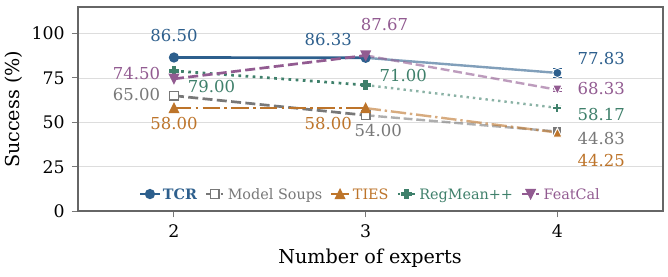}
\caption{Success of TCR, Model Soups, TIES, RegMean++ (with spectral smoothing), and FeatCal for Spatial+Goal and Spatial+Object+Goal subsets (one evaluation each; no error bars), followed by the four-expert main-table endpoint (mean $\pm$ sample standard deviation over three evaluations). The four-expert TCR checkpoints differ from those of the smaller subsets. Each evaluation contributes 100 episodes per included suite. Task coverage grows with the expert set; this is not a fixed-task scaling comparison.}
\label{fig:goal_first_expert_count}
\end{figure}

\paragraph{Comparison with other mergers on nested subsets.}
\label{app:expert_count}
We use three fixed nested expert sets: Spatial\allowbreak{}+Goal, Spatial\allowbreak{}+Object\allowbreak{}+Goal, and Spatial\allowbreak{}+Object\allowbreak{}+Goal\allowbreak{}+Long.
The two- and three-expert subset policies are each evaluated once on their included suites, using 100 episodes per suite from the fixed first-repeat selection.
Model Soups achieves 130/200 (65.00\%) and 162/300 (54.00\%); TCR achieves 173/200 (86.50\%) and 259/300 (86.33\%). TIES achieves 116/200 (58.00\%) and 174/300 (58.00\%). RegMean++ achieves 158/200 (79.00\%) and 213/300 (71.00\%). FeatCal achieves 149/200 (74.50\%) and 263/300 (87.67\%).
The four-expert endpoint reuses the main-table comparison over three 400-episode evaluations: Model Soups achieves 538/1,200 (44.83$\pm$1.81\%), TCR achieves 934/1,200 (77.83$\pm$2.52\%), TIES achieves 531/1,200 (44.25$\pm$1.30\%), RegMean++ achieves 698/1,200 (58.17$\pm$1.04\%), and FeatCal achieves 820/1,200 (68.33$\pm$0.76\%) (Figure~\ref{fig:goal_first_expert_count}).
TCR leads the compared methods at two and four experts, whereas FeatCal is 1.33 percentage points higher at three experts (four additional successful episodes).
Thus, the margin over direct averaging does not imply a consistent advantage over every functional merger as the expert set changes.
The four-expert TCR checkpoints and number of evaluations therefore differ from the smaller fixed-subset points; the task set also expands with expert count. The connected points are descriptive and do not isolate an expert-count effect.

\subsection{What Characterizes Mergeability}
\label{app:limits_fidelity}

\begin{figure}[t]
\centering
\includegraphics[width=\linewidth]{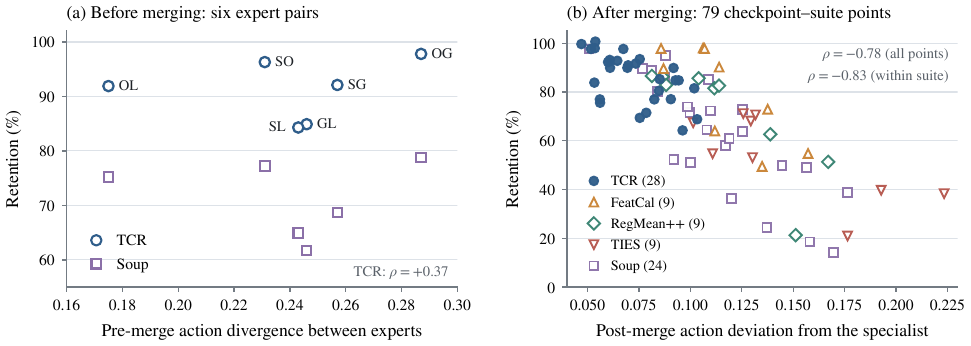}
\caption{(a) Pre-merge action divergence between two specialists versus retention of TCR and Model Soups for the six expert pairs; the Spearman correlation refers to TCR. (b) Post-merge action deviation of the merged model from the suite specialist versus retention for all 79 checkpoint--suite points of the five merging methods; counts in parentheses are points per method.}
\label{fig:limits_fidelity}
\end{figure}

\paragraph{Pre-merge divergence.}
For each of the six expert pairs, we compute three pre-merge measures: the action divergence between the two specialists on 24 recorded expert-execution requests per suite (mean absolute error of the first 10 end-effector position steps in environment units; the same specialist under different noise gives about 0.024), the cosine similarity of their task vectors, and their parameter distance. None shows a reliable association with pairwise TCR retention: the Spearman correlations are $+0.37$, $+0.14$, and $-0.09$, respectively ($+0.09$ for the cosine restricted to the action expert). Object--Goal has the largest action divergence (0.287) but the highest retention (97.8\%), and all task-vector cosines lie in a narrow range (0.051--0.063). With six pairs, these results show only the absence of an observed predictive relation (Figure~\ref{fig:limits_fidelity}a). Representations built from how models respond to shared inputs, such as the attribution-based embeddings of ABLE for language models~\citep{wang-etal-2026-able}, offer another way to relate models; whether such measures predict the mergeability of robot policies remains untested.

\paragraph{Post-merge fidelity.}
We apply the same offline protocol to the merged models, measuring each merged model's action deviation from the suite specialist on that specialist's execution requests. The analysis covers 79 checkpoint--suite points from 30 merged checkpoints: all 11 TCR subsets, 10 Model Soups subsets, and the three Spatial$\to$Goal subsets of FeatCal, RegMean++, and TIES. The deviation is strongly associated with retention (Figure~\ref{fig:limits_fidelity}b): $\rho=-0.78$ over all points and $-0.83$ within suites. The association is not specific to TCR or Model Soups; for FeatCal, RegMean++, and TIES together, $\rho=-0.80$. On the Spatial$\to$Goal chain, the five methods have the same ordering by mean deviation as by mean retention, with TCR lowest (0.086, versus 0.119 for FeatCal, 0.120 for RegMean++, 0.132 for Model Soups, and 0.151 for TIES). This is an observed association, not a causal law; the measure requires a merged model and therefore serves as an offline diagnostic rather than a pre-merge predictor. It complements Section~\ref{sec:mechanism}, where the ranking of nine four-expert merges by action MSE correlates less with success on Long ($\rho=0.52$) than on the other suites ($\rho=0.95$): across expert sets and methods, fidelity tracks retention, but long-horizon success also depends on recovery from states that the offline requests do not cover.

\paragraph{Shared-mapping residual.}
To relate the measurements to Proposition~1, we fit, for each of the 418 calibrated modules, the best shared linear map for an expert set by ridge regression centered at the expert mean, using the features each expert's execution produces under merged-prefix replay, and evaluate it on held-out requests (the modules have more input dimensions than rows per expert, so an in-sample residual would be zero). This held-out shared-mapping residual increases with the number of experts in every suite, and with four experts its ordering across suites (Object $<$ Spatial $<$ Goal $<$ Long) matches the ordering of retention. The residual also contains finite-sample, regularization, and generalization error, so it is not a measurement of the lower bound in Eq.~\eqref{eq:conflict_bound}; within a suite, it moves together with the number of experts and therefore does not separate conflict from expert count.

To separate the cost of sharing from estimation error, we split each expert's requests into fitting, selection, and test thirds, choose the ridge strength on the selection third, and compare the shared map with independent maps fitted to each expert under the same split and regularization. Averaged over the four suites, the shared residual grows by factors of 1.36 and 1.57 from two to three and four experts, and the excess of the shared over the independent residual grows by factors of 1.39 and 1.64; the excess increases with the number of experts in every suite. The independent residual also grows (1.28 and 1.41), because its ridge term is centered at the mean of the experts in the set, and it is at least as strongly associated with retention across the 28 checkpoint--suite points ($\rho=-0.90$, versus $-0.80$ for the excess). The association with retention is therefore not specific to sharing, and we report only that the excess of a shared map over independent maps increases with the number of experts.

\paragraph{Relation to the limitations.}
Section~\ref{sec:discussion} summarizes these bounds and discusses three directions for future work, which the findings above connect to as follows. The growth of the shared-mapping residual with the number of experts motivates continual and grouped merging that controls interference as specialists are added. The requirement of compatible architectures and observation and action interfaces motivates black-box and cross-embodiment merging through representation or interface alignment. Since post-merge fidelity on execution inputs tracks retention while pre-merge similarity does not, offline fidelity checks can screen merges before closed-loop rollouts, and recovery-aware calibration can target the states that expert rollouts do not cover.

\end{document}